%% file: sample-manuscript.tex
\documentclass[manuscript,screen]{acmart}
\usepackage{lineno}
\usepackage{microtype}
\usepackage{enumitem}
\usepackage{graphicx}
\usepackage{pifont}
\usepackage{amsfonts}
\usepackage{amsmath}
\usepackage{booktabs}
\usepackage{longtable}
\usepackage[table]{xcolor}
\usepackage{ragged2e}
\usepackage{array}
\usepackage{makecell}
\usepackage{multirow}
\usepackage{tabularx}
\usepackage{subcaption}
\usepackage{algpseudocode}
\usepackage{algorithm} 
\usepackage[most,skins,theorems]{tcolorbox}
\usepackage{wrapfig}

\usepackage{pgf}
\usepackage{tikz} 
\usepackage[edges]{forest}
\definecolor{customred}{RGB}{207, 55, 49}
\definecolor{customorange}{RGB}{255, 128, 0}
\definecolor{customgreen}{RGB}{0, 204, 0}
\definecolor{customyellow}{RGB}{251, 239, 214}

\definecolor{myblue}{rgb}{0.9, 0.1, 0.94}
\definecolor{mygreen}{rgb}{0.64, 0.56, 0.88}
\definecolor{myyellow}{rgb}{0.68, 0.6, 0.1}
\definecolor{fancygreen}{rgb}{0.33, 0.68, 0.20}
\definecolor{salmon}{rgb}{0.94, 0.52, 0.49}
\definecolor{tablegreen}{rgb}{0.82, 0.94, 0.75}
\definecolor{tableblue}{rgb}{0.81, 0.90, 0.94}
\definecolor{tablered}{rgb}{0.97, 0.85, 0.85}
\definecolor{tableorange}{rgb}{0.96, 0.85, 0.81}
\definecolor{myorange}{rgb}{1.0, 0.49, 0.0}	
\definecolor{tlgreen}{rgb}{0.33, 0.68, 0.20}

\usepackage{float}

\newcommand{\cmark}{\ding{51}}
\newcommand{\xmark}{\ding{55}}

\newcolumntype{Y}{>{\Centering\arraybackslash}m{3.8cm}}  
\newcolumntype{L}{>{\RaggedRight\arraybackslash}X}       

\newtcolorbox{myboxi}[1][]{
  breakable,
  title=#1,
  colback=blue!5,
  colbacktitle=blue!5,
  coltitle=black,
  fonttitle=\bfseries,
  bottomrule=1pt,
  toprule=1pt,
  leftrule=1pt,
  rightrule=1pt,
  titlerule=0pt,
  arc=0pt,
  outer arc=0pt,
  colframe=blue,
}

\newtcolorbox{myboxiii}[1][]{
  breakable,
  title=#1,
  colback=red!5,
  colbacktitle=red!5,
  coltitle=black,
  fonttitle=\bfseries,
  bottomrule=0pt,
  toprule=0pt,
  leftrule=2pt,
  rightrule=2pt,
  titlerule=0pt,
  arc=0pt,
  outer arc=0pt,
  colframe=red,
}

\newtcolorbox{myboxnote}[1][]{
  breakable,
  title=#1,
  colback=orange!0,
  colbacktitle=orange!0,
  coltitle=black,
  fonttitle=\bfseries,
  bottomrule=0pt,
  toprule=0pt,
  leftrule=2pt,
  rightrule=2pt,
  titlerule=0pt,
  arc=0pt,
  outer arc=0pt,
  colframe=orange,
}

\usepackage{epigraph}

\newtcolorbox{myboxii}[1][]{
  breakable,
  freelance,
  title=#1,
  colback=white,
  colbacktitle=white,
  coltitle=black,
  fonttitle=\bfseries,
  bottomrule=0pt,
  boxrule=0pt,
  colframe=white,
  overlay unbroken and first={
  \draw[red!75!black,line width=3pt]
    ([xshift=5pt]frame.north west) -- 
    (frame.north west) -- 
    (frame.south west);
  \draw[red!75!black,line width=3pt]
    ([xshift=-5pt]frame.north east) -- 
    (frame.north east) -- 
    (frame.south east);
  },
  overlay unbroken app={
  \draw[red!75!black,line width=3pt,line cap=rect]
    (frame.south west) -- 
    ([xshift=5pt]frame.south west);
  \draw[red!75!black,line width=3pt,line cap=rect]
    (frame.south east) -- 
    ([xshift=-5pt]frame.south east);
  },
  overlay middle and last={
  \draw[red!75!black,line width=3pt]
    (frame.north west) -- 
    (frame.south west);
  \draw[red!75!black,line width=3pt]
    (frame.north east) -- 
    (frame.south east);
  },
  overlay last app={
  \draw[red!75!black,line width=3pt,line cap=rect]
    (frame.south west) --
    ([xshift=5pt]frame.south west);
  \draw[red!75!black,line width=3pt,line cap=rect]
    (frame.south east) --
    ([xshift=-5pt]frame.south east);
  },
}

\AtBeginDocument{%
  }

\setcopyright{acmlicensed}
\copyrightyear{2025}
\acmYear{2025}
\acmDOI{XXXXXXX.XXXXXXX}

\acmJournal{JACM}
\acmVolume{37}
\acmNumber{4}
\acmArticle{111}
\acmMonth{10}

\begin{document}

\title{A Comprehensive Survey on Reinforcement Learning-based Agentic Search: Foundations, Roles, Optimizations, Evaluations, and Applications}



\author{Minhua Lin}
\email{mfl5681@psu.edu}
\author{Zongyu Wu}
\affiliation{%
  \institution{The Pennsylvania State University}
  \city{University Park}
  \country{USA}
}


\author{Zhichao Xu}
\affiliation{%
  \institution{The University of Utah}
  \city{Salt Lake City}
  \country{USA}
}

\author{Hui Liu}
\author{Xianfeng Tang}
\affiliation{%
  \institution{Amazon}
  \country{USA}
}



\author{Qi He}
\affiliation{%
  \institution{Microsoft}
  \country{USA}
}

\author{Charu Aggarwal}
\affiliation{%
  \institution{IBM T.J. Watson Research Center}
  \country{USA}
}

\author{Hui Liu}
\affiliation{%
  \institution{Michigan State University}
  \country{USA}
}

\author{Xiang Zhang}
\affiliation{%
  \institution{The Pennsylvania State University}
  \city{University Park}
  \country{USA}
}

\author{Suhang Wang}
\authornote{Corresponding Author}
\affiliation{%
  \institution{The Pennsylvania State University}
  \city{University Park}
  \country{USA}
}
\email{szw494@psu.edu}

\renewcommand{\shortauthors}{Minhua Lin et al.}

\input{0_Abstract_v2}

\ccsdesc[500]{Computing methodologies~Artificial intelligence}
\keywords{Large Language Models, Reinforcement Learning, Agentic Search}

\received{20 February 2007}
\received[revised]{12 March 2009}
\received[accepted]{5 June 2009}

\maketitle
\input{1_Intro_v4}

\input{2_Background_v4}
\input{5_What_is_for_v3}

\input{4_How_to_use_v3}

\input{6_Where_to_use_v2}
\input{7_Evaluation}
\input{8_Future_direction_v2}
\input{9_Conclusion}
\bibliographystyle{ACM-Reference-Format}
\bibliography{custom}


\end{document}

%% file: 0_Abstract_v2.tex
\begin{abstract}
The advent of large language models (LLMs) has transformed information access and reasoning through open-ended natural language interaction.
However, LLMs remain limited by static knowledge, factual hallucinations, and the inability to retrieve real-time or domain-specific information.
Retrieval-Augmented Generation (RAG) mitigates these issues by grounding model outputs in external evidence, but traditional RAG pipelines are often single turn and heuristic, lacking adaptive control over retrieval and reasoning.
Recent advances in \emph{agentic search} address these limitations by enabling LLMs to plan, retrieve, and reflect through multi-step interaction with search environments.
Within this paradigm, reinforcement learning (RL) offers a powerful mechanism for adaptive and self-improving search behavior.
This survey provides the first comprehensive overview of \emph{RL-based agentic search}, organizing the emerging field along three complementary dimensions: (i) \emph{What RL is for} (functional roles), (ii) \emph{How RL is used} (optimization strategies), and (iii) \emph{Where RL is applied} (scope of optimization). We summarize representative methods, evaluation protocols, and applications, and discuss open challenges and future directions toward building reliable and scalable RL driven agentic search systems. We hope this survey will inspire future research on the integration of RL and agentic search. Our repository is available at \url{https://github.com/ventr1c/Awesome-RL-based-Agentic-Search-Papers}.

\end{abstract}

%% file: 1_Intro_v4.tex
\section{Introduction}
\label{sec:introduction}

Large Language Models (LLMs)~\cite{ouyang2022training,touvron2023llama,zhang2025agent} have shown unprecedented capabilities in natural language understanding, reasoning, and generation, fundamentally reshaping how users access and interact with information. Despite these advantages, LLMs still suffer from several limitations: they are constrained by static knowledge cutoffs~\cite{cheng2024dated}, prone to factual hallucinations~\citep{sahoo2024comprehensivehallusurvey}, and unable to access real-time or domain-specific information. To address these challenges, the paradigm of \emph{Retrieval-Augmented Generation (RAG)}~\cite{lewis2020retrieval,gao2023retrieval} has emerged as a popular solution. RAG combines the reasoning power of LLMs with the precision of classical information retrieval (IR) techniques such as TF–IDF~\cite{sparck1972statistical,aizawa2003information}, BM25~\cite{robertson1995okapi,robertson2009probabilistic}, and link-analysis models like PageRank~\cite{brin1998anatomy,page1999pagerank,bianchini2005inside}. By retrieving evidence from external knowledge bases and conditioning responses on this context, 
RAG enables LLMs to generate more accurate and factually grounded outputs, particularly in knowledge-intensive tasks~\cite{asai2024reliable,borgeaud22ragenhance,fan2024ragsurvey}. 

However, traditional RAG systems~\cite{chen2017reading} are typically single-turn and heuristic-driven: they retrieve once and generate once, lacking the ability to iteratively refine queries or adapt retrieval strategies based on intermediate feedback. Retrieved documents may be irrelevant or noisy, hindering downstream reasoning~\cite{jiang2023activerag,chang-etal-2025-main,jin2025longcontext,jin2025searchr1}. Moreover, LLMs often struggle to fully utilize retrieved evidence, limiting the overall effectiveness of the pipeline.
These limitations motivates the development of more \emph{agentic search systems}, where LLMs act as autonomous decision-makers that dynamically plan, retrieve, reason, and reflect over multiple steps.


To this end, researchers have proposed \emph{search agents}
i.e., LLM-based systems capable of multi-step interaction with search environments~\cite{jiang2024rag,zheng2025deepresearcher}. Unlike traditional RAG, search agents can iteratively issue and refine queries, assess the quality of retrieved results, and dynamically adapt their strategies to solve complex, multi-hop tasks. This shift from passive retrieval to active agency represents a paradigm change in information-seeking. However, early search agents often heavily rely on handcrafted prompting~\cite{li2025search} or supervised fine-tuning~\cite{qin2023toolllm,asai2024self}, limiting their ability to autonomously discover optimal strategies.  

Recently, reinforcement learning (RL)~\cite{sutton1998reinforcement} has emerged as a promising paradigm for developing adaptive and autonomous search agents~\cite{jin2025searchr1, wang2025stepsearch}. We define \emph{RL-based agentic search} as 
training an LLM as a decision-making agent that interacts with a search environment, receives external feedback, and iteratively improves its strategy to maximize rewards.
This formulation highlights three key aspects: (i) \emph{autonomy}, where the agent determines its search actions; (ii) \emph{learning}, where strategies are acquired through reinforcement rather than manual design; and (iii) \emph{interaction}, where the agent engages in multi-turn exchanges with search environments to refine reasoning and retrieval.

\begin{figure}[!htbp]
\centering
\begin{spacing}{1.}
\resizebox{0.95\linewidth}{!}{
\begin{forest}
	for tree={
		draw,
		shape=rectangle,
		rounded corners,
		top color=white,
		grow'=0,
		l sep'=1.2em,
		reversed=true,
		anchor=west,
		child anchor=west,    
	},
	forked edges,
	root/.style={
    draw=customred,
		rotate=90, shading angle=90, bottom color=white!40,
		anchor=north, font=\normalsize, inner sep=0.5em},
	level1/.style={
    draw=customred,
            text width=2cm,
		bottom color=white!30, font=\normalsize, inner sep=0.3em,
		s sep=0.2em},
	level2/.style={
            draw=customred,
            text width=4cm,
		bottom color=white!40, font=\small, inner sep=0.25em, 
		s sep=0.1em},
	level3/.style={
            draw=customred,
            text width=4cm,  
		bottom color=white!40, font=\small, inner sep=0.2em, 
		l sep'=0.5em},
	level4/.style={
            draw=customred,
            text width=6.7cm,
            top color=customyellow!100, 
		bottom color=customyellow!100, font=\small, inner sep=0.2em,
		l sep'=0.5em},
	where n=0{root}{},
	where level=1{level1}{},
	where level=2{level2}{},
	where level=3{level3}{},
	where level>=4{level4}{},
	[Total Name
	    [Introduction (\S \ref{sec:introduction})]
		[Background (\S \ref{sec:background})
                [LLMs as Agents (\S \ref{sec:llm_as_agents})
                ]
                [From IR to Agentic Search (\S \ref{sec:from_ir_to_search})
                    [Traditional IR (\S \ref{sec:tradition_ir})
                        [
                            TF-IDF~\cite{aizawa2003information}; BM25~\cite{robertson2009probabilistic}; PageRank~\cite{bianchini2005inside}
                        ]
                    ]
                    [RAG (\S \ref{sec:rag})
                        [Naive RAG~\cite{lewis2020retrieval}; Iterative RAG~\cite{trivedi2022interleaving,asai2024self}]
                    ]
                    [Agentic Search (\S \ref{sec:agentic_search})
                    ]
                ]
                [RL Basic (\S \ref{sec:rl_basic})
                    [On-policy Optimization (\S \ref{sec:tradition_ir})
                        [
                            PPO~\cite{schulman2017proximal}; GRPO~\cite{shao2024deepseekmath};
                            DAPO~\cite{yu2025dapo}
                        ]
                    ]
                    [Off-policy Optimization (\S \ref{sec:rag})
                        [
                            DPO~\cite{rafailov2023direct}; ReMix~\cite{liang2025squeeze}
                        ]
                    ]
                ]
                [RL-based Agentic Search (\S \ref{sec:rl-based_agentic_search})]
		]
        [What RL is for (\S \ref{sec:what_RL_is_for})
            [Retrieval Control (\S \ref{sec:retrieval_control})          
                [Adaptive Search Decisions (\S \ref{sec:adaptive search decision})
                   [
                       Search-R1~\cite{jin2025searchr1}; ReSearch~\cite{chen2025learning}; R1-Searcher~\cite{song2025r1}; DeepRAG~\cite{guan2025deeprag}; UR$^2$~\cite{li2025ur}; SSRL~\cite{fan2025ssrl}; VERITAS~\cite{xu2025VERITAS}
                   ]
                ]
                [Search Intensity (\S \ref{sec:search_intensity})
                     [Pangu DeepDiver~\cite{shi2025pangu}; ReZero~\cite{dao2025rezero}; StepSearch~\cite{wang2025stepsearch}; ReasonRAG~\cite{zhang2025process}; WebSailor-V2~\cite{li2025websailorv2}
                     ]
                ]
                [Search Efficiency (\S \ref{sec:search_efficiency})
                [IKEA~\cite{huang2025reinforced}; R1-Searcher++~\cite{song2025r1++}; 
                    DeepRAG~\cite{guan2025deeprag}; 
                    Search Wisely~\cite{wu2025search}; StepSearch~\cite{wang2025stepsearch}; ZeroSearch~\cite{sun2025zerosearch}; ParallelSearch~\cite{zhao2025parallelsearch}; RAG-R1~\cite{tan2025ragr1};
                    WebThinker~\cite{li2025webthinker}...
                    ]
                ]
            ]
            [Query Optimization (\S \ref{sec:query_optimization})      
                [Conversational Reformulation  (\S \ref{sec:conver_reformulation})
                    [
                    ConvSearch-R1~\cite{zhu2025convsearch}; MaskSearch~\cite{wu2025masksearch}; RAG-R1~\cite{tan2025ragr1}; ParallelSearch~\cite{zhao2025parallelsearch}
                    ]
                ]
                [Retriever-aware Optimization (\S \ref{sec:retriever_optimization})
                [
                DeepRetrieval~\cite{jiang2025deepretrieval}; ZeroSearch~\cite{sun2025zerosearch}; s3~\cite{jiang2025s3}; WebThinker~\cite{li2025webthinker}
                ]
                ]
            ]
            [Reasoning-Retrieval Integration (\S \ref{sec:reasoning-retrieval-integration})      
                [Reasoning–Search Interleaving (\S \ref{sec:reasoning_search_interleaving})
                 [R-Search~\cite{zhao2025r}; AutoRefine~\cite{shi2025search}; EvolveSearch~\cite{zhang2025evolvesearch}; ReasonRAG~\cite{zhang2025process}; O${^2}$-Searcher~\cite{mei2025o2} ...]
                ]
                [Context and Memory Management (\S \ref{sec:context_memory_management})
                [ReSum~\cite{wu2025resum}; SFR-DeepResearch~\cite{nguyen2025sfr}]
                ]
            ]
            [Multi-Agent Collaboration (\S \ref{sec:what_is_for_multi_agent_collaboration})      
                [Planner–Executor Architectures (\S \ref{sec:planner_executor_architectures})
                [
                MAO-ARAG~\cite{chen2025mao}; OPERA~\cite{liu2025opera}; AI-SearchPlanner~\cite{mei2025ai}
                ]
                ]
                [Cooperative Multi-Agent Systems (\S \ref{sec:cooperative_multi-agent_system})
                [SIRAG~\cite{wang2025sirag}; MMOA-RAG~\cite{chen2025improving}; AgentGym-RL~\cite{xi2025agentgym}; Chain-of-Agents~\cite{li2025chain}]
                ]
            ]
            [Multi-Agent Collaboration (\S \ref{sec:tool_knowledge_integration})      
                [Multi-tool and Multi-modality Reasoning (\S \ref{sec:multi_tool_multi_modal_reasoning})
                [Tool-Star~\cite{dong2025tool}; VerlTool~\cite{jiang2025verltool}; WebWatcher~\cite{geng2025webwatcher}; AI-SearchPlanner~\cite{mei2025ai}; WebSailor-V2~\cite{li2025websailorv2}; Visual-ARFT~\cite{liu2025visual}; VRAG-RL~\cite{wang2025vrag}; MMSearch-R1~\cite{wu2025mmsearch}]
                ]
                [Structured Knowledge Navigation (\S \ref{sec:structure_knowledge_navigatino})
                [GRAIL~\cite{chang2025grail}; DynaSearcher~\cite{hao2025dynasearcher}]]
            ]
        ]
        [How RL is Used (\S \ref{sec:how_RL_is_used})
            [Training Regime (\S \ref{sec:training_regime})          
            ]
            [Reward Design (\S \ref{sec:reward_design}))      
                [Outcome-level Rewards (\S \ref{sec:outcome_reward})
                    [Answer Correctness~\cite{jin2025searchr1,song2025r1,sun2025simpledeepsearcher}; Format Reward~\cite{chen2025learning}; Search Efficiency~\cite{song2025r1++,guan2025deeprag,li2025ur}; Search Effectiveness~\cite{shi2025pangu,dao2025rezero,mei2025ai,jiang2025s3}; Tool Effectiveness~\cite{dao2025lucy}]
                ]
                [Process-level Rewards (\S \ref{sec:process_reward})
                    [Trajectory Quality~\cite{shi2025iterative,goldie2025synthetic,zhang2025process};
                    Intermediate Action Quality~\cite{wang2025stepsearch,wang2025sirag,zhao2025parallelsearch,zhu2025convsearch,liu2025visual};
                    Retrieval Quality~\cite{wang2025stepsearch,zhao2025r,hu2025coordinating,yu2025medresearcher}
                    ]
                ]
            ]
        ]
        [Where RL is Applied (\S \ref{sec:where_RL_is_applied})
            [Agent-level (\S \ref{sec:agent_level_scope})          
                [Single-agent Optimization (\S \ref{sec:single_agent})
                    [Search-R1\cite{jin2025searchr1}; ReSearch\cite{chen2025learning}; R1-Searcher++\cite{song2025r1++}; AutoCoA~\cite{zhang2025agent}; DeepRAG\cite{guan2025deeprag}; WebSailor\cite{li2025websailor} ...]
                ]
                [Multi-agent Coordination (\S \ref{sec:multi_agent_corrdination})
                    [HARIS~\cite{hu2025coordinating}; SIRAG\cite{wang2025sirag}; MAO-ARAG\cite{chen2025mao}; MMOA-RAG\cite{chen2025improving}; OPERA~\cite{liu2025opera}]
                ]
            ]
            [Module-level \& Step-level (\S \ref{sec:module_level_step_level}))      
                [Module-level Optimization (\S \ref{sec:module_level_opt})
                    [s3\cite{jiang2025s3}; AI-SearchPlanner\cite{mei2025ai}; DeepResearcher~\cite{zheng2025deepresearcher}]
                ]
                [Step-level Optimization (\S \ref{sec:step_level_opt})
                    [StepSearch\cite{wang2025stepsearch}; AutoRefine\cite{shi2025search}; Search Wisely\cite{wu2025search}; ConvSearch-R1~\cite{zhu2025convsearch}; Atom-Searcher~\cite{deng2025atom}; ReasonRAG\cite{zhang2025process}; SWiRL~\cite{goldie2025synthetic}]
                ]
            ]
            [System-level (\S \ref{sec:system_level}))      
                [Unified RL-based Framework (\S \ref{sec:unified_RL-based_framework})
                    [AgentGym-RL\cite{xi2025agentgym}; Verl~\cite{sheng2024hybridflow}; VerlTool\cite{jiang2025verltool}; RAG-Gym\cite{xiong2025rag}; Chain-of-Agents\cite{li2025chain}]
                ]
            ]
        ]
        [Evaluation and Application (\S \ref{sec:evaluation_application})
            [Metrics (\S \ref{sec:metric}))      
                [Answer Quality (\S \ref{sec:answer_quality})
                    [EM~\cite{jin2025searchr1}; F1 score~\cite{chen2025learning}; LLM Judge~\cite{hao2025dynasearcher} ...]
                ]
                [Search Effectiveness (\S \ref{sec:search_effectiveness})
                    [Recall; MRR; NDCG~\cite{zhu2025convsearch,jiang2025deepretrieval} ...]
                ]
                [Search Efficiency (\S \ref{sec:metric_search_efficiency})
                    [Query Number~\cite{shi2025pangu}; API Call Cost~\cite{chen2025mao}; Response Time~\cite{liu2025sefrqo}; Search Redundancy~\cite{song2025r1++}]
                ]
                [Specialized Process Metric (\S \ref{sec:process_metric})
                    [Query Quality~\cite{wang2025sirag}; Evidence Utilization Rate~\cite{zhao2025r}  ...]
                ]
            ]
            [Applications (\S \ref{sec:applications}))]
            [Dataset (\S \ref{sec:datasets})          
                [Knowledge-Intensive QA (\S \ref{sec:knowledge_intensive_qa})
                    [NQ~\cite{kwiatkowski2019natural};
TriviaQA~\cite{joshi2017triviaqa}; HotpotQA~\cite{yang2018hotpotqa}; 2WikiMultiHopQA~\cite{ho2020constructing}; MuSiQue~\cite{trivedi2022musique}; PopQA~\cite{mallen2022not}; CAG~\cite{pan2024cag}; C-SimpleQA~\cite{wei2024measuring}; 
SuperGPQA~\cite{du2025supergpqa};
FEVER~\cite{thorne2018fever} ...
]
                ]
                [Web-based Search (\S \ref{sec:web_based_search_bench})
                    [Mind2Web~\cite{gou2025mind2web}; 
WebArena~\cite{zhou2023webarena};
WebWalkerQA~\cite{wu2025webwalker};
AgentBench~\cite{liu2024agentbench};
BrowseComp-en~\cite{bc_en} ...]
                ]
                [Knowledge Sources (\S \ref{sec:knowledge_sources})
                    [wiki-dump~\cite{wikimedia_dumps}; Common Crawl~\cite{commoncrawl_overview}; KILT~\cite{petroni2020kilt}; PubMed~\cite{pubmed_about}; Arxiv~\cite{arxiv_about}]
                ]
                [Multi-modal (\S \ref{sec:dataset_multi_modal_search})
                    [InfoSeek~\cite{chen2023can}; MMSearch~\cite{jiang2024mmsearch}; MMSearch-Plus~\cite{tao2025mmsearch} SimpleVQA~\cite{cheng2025simplevqa}; LiveVQA~\cite{fu2025livevqa}; MM-BrowseComp~\cite{li2025mm}; MAT-Search~\cite{liu2025visual}; Mocheg~\cite{yao2023end}; MFC-Bench~\cite{wang2024mfc}]
                ]
                [Conversational and Multi-turn Search (\S \ref{sec:conversational})
                    [CoQA~\cite{reddy2019coqa}; QuAC~\cite{choi2018quac}; MSMarco~\cite{bajaj2016ms}; TopiOCQA~\cite{adlakha2022topiocqa}; QReCC~\cite{anantha2021open}; OR-QuAC~\cite{qu2020open}; NarrativeQA~\cite{kovcisky2018narrativeqa} ...]
                ]
                [Domain-specific Search (\S \ref{sec:domain_specific})
                    [MATH~\cite{hendrycks2021measuring}; MedQA~\cite{jin2021disease};
OlympiadBench~\cite{he2024olympiadbench}; HLE~\cite{phan2025humanity}; MIRAGE~\cite{dongre2025mirage}; HERB~\cite{choubey2025deepsearch};
SciQ~\cite{welbl2017crowdsourcing} ...]
                ]
            ]
        ]
        [Challenges and Future Direction (\S \ref{sec:challenge_future_direction})]
	]
\end{forest}
}
\end{spacing}
\vspace{-0.75em}
\caption{Overview of RL-based Agentic Search.}
\label{fig:overview}
\end{figure}
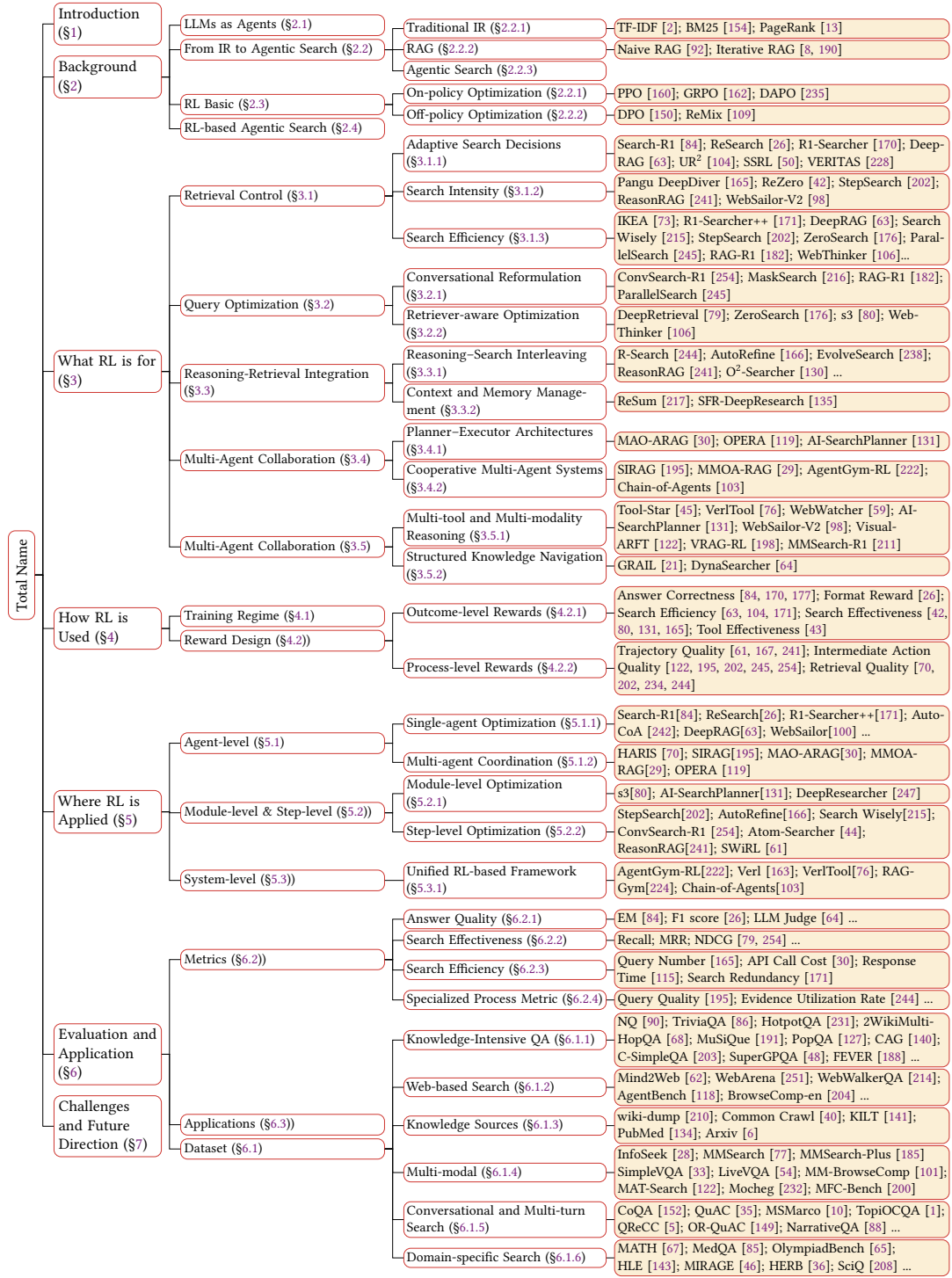

Despite rapid progress, a systematic understanding of RL–based agentic search remains limited. As summarized in Table~\ref{tab:survey_comparison}, recent surveys~\cite{xi2025survey, li2025reinforcement, gao2025synergizing} have examined agentic search from various perspectives. However, they either pay less attention to RL~\cite{xi2025survey} or focus on specific sub-domains such as Deep Research~\cite{li2025reinforcement} and RAG~\cite{gao2025synergizing}. The role of RL in enabling adaptive and autonomous search behaviors remains underexplored.
In contrast, this paper presents the \emph{first} comprehensive survey dedicated to RL-based agentic search, aiming to clarify how RL benefits agentic search across three complementary dimensions:
(i) \emph{What RL is for}, describing its functional roles in guiding retrieval, reasoning, and decision making; (ii) \emph{How RL is used}, covering optimization strategies such as reward design, policy learning, and advanced training methods; and (iii) \emph{Where RL is applied}, examining the scope of RL intervention from the agent level to the step and module levels. For each dimension, we review representative methods and summarize emerging trends. The overview structure of our paper is shown in Figure~\ref{fig:overview}.

This paper is organized as follows: Section~\ref{sec:background} introduces the foundations of agentic search and RL. From Sections~\ref{sec:what_RL_is_for} to~\ref{sec:where_RL_is_applied}, we examine RL for agentic search from the three perspectives outlined above. Section~\ref{sec:evaluation_application} reviews evaluation metrics and representative applications, and Sec.~\ref{sec:challenge_future_direction} concludes with open challenges and future directions.

\begin{table*}[t]
\centering
\small
\setlength{\tabcolsep}{4pt}
\caption{Comparison of representative surveys and this work. 
\cmark\ indicates the topic is a primary focus; \xmark\ indicates limited or no coverage. 
Unlike prior surveys that focus on non-RL agentic RAG or general search agents, or on RL methods limited to building deep-research systems, 
our work uniquely unifies \textbf{RL foundations} with \textbf{agentic search behavior}, 
analyzing how RL benefits agentic search, how it optimizes search agents, and how such systems can be effectively evaluated.}
\label{tab:survey_comparison}
\vskip -0.6em
\begin{tabularx}{\textwidth}{
>{\RaggedRight\arraybackslash}m{2.2cm}
>{\RaggedRight\arraybackslash}m{3.1cm}
>{\centering\arraybackslash}m{1.3cm}
>{\centering\arraybackslash}m{1.5cm}
>{\centering\arraybackslash}m{1.6cm}
>{\centering\arraybackslash}m{1.4cm}
>{\RaggedRight\arraybackslash}m{3cm}
}
\toprule
\textbf{Survey} & \textbf{Analytical Focus} & \textbf{RL Foundations} & \textbf{Search Behavior} & \textbf{Reasoning Integration} & \textbf{Evaluation Scope} & \textbf{Application Scope} \\
\midrule
\citet{singh2025agentic}      
& Agentic RAG                     
& \xmark & \xmark & \cmark & \cmark 
&  \cmark\\

\citet{liang2025agenticsurvey}
& Reasoning in RAG           
& \xmark & \xmark & \cmark & \xmark 
& \xmark  \\

\citet{gao2025synergizing}
& Reasoning in RAG          
& \xmark & \xmark & \cmark & \cmark 
& \cmark \\

\citet{xi2025survey}          
& General search agents            
& \xmark & \cmark & \xmark & \cmark 
& \cmark \\

\citet{li2025reinforcement}   
& RL-based deep research
& \cmark & \xmark & \xmark & \cmark 
& \cmark (Deep Research)\\

\textbf{Ours} 
& \textbf{RL-based agentic search} 
& \textbf{\cmark} & \textbf{\cmark} & \textbf{\cmark} & \textbf{\cmark} 
& \cmark \\
\bottomrule
\end{tabularx}
\end{table*}

%% file: 2_Background_v4.tex
\section{Background and Preliminary}
\label{sec:background}
\subsection{Large Language Models as Agents}
\label{sec:llm_as_agents}
LLMs~\cite{ouyang2022training,touvron2023llama, yang2025qwen3,wang2025survey,lin2025far,zhang2024offline} have demonstrated remarkable capabilities in text understanding, reasoning, and generation, fundamentally reshaping how humans access and interact with information. Their success has enabled natural language interfaces to diverse knowledge resources. However, these models remain limited by static training corpora, hallucinations, and their inability to access real-time or domain-specific knowledge directly~\cite{ji2023survey}. 
To overcome these, researchers have increasingly augmented LLMs with external information sources and decision-making capabilities. A prominent direction is \emph{Retrieval-Augmented Generation (RAG)}~\cite{lewis2020retrieval,gao2023retrieval,liu2025exposing}, where LLMs query external knowledge bases to ground responses in retrieved evidence. Building on this paradigm, recent advances~\cite{qin2023toolllm, zheng2025deepresearcher} further position LLMs as \emph{agentic systems}, capable of invoking external tools such as \emph{search engines, code interpreters, knowledge-base query APIs, and web browsers} to interact with dynamic environments and perform multi-step reasoning.

\subsection{From Traditional IR to Agentic Search}
\label{sec:from_ir_to_search}
\subsubsection{Traditional IR}
\label{sec:tradition_ir}
In classical information retrieval (IR), the primary objective is to return a ranked list of documents that best match a user query, relying on statistical models such as TF–IDF~\cite{sparck1972statistical} and BM25~\cite{robertson1995okapi}, as well as link analysis methods like PageRank~\cite{brin1998anatomy,page1999pagerank} that incorporates metadata beyond pure texts. Retrieval itself is the endpoint of the process, leaving users to interpret and synthesize the results. In addition, while effective for many tasks, traditional IR methods are fundamentally limited in their ability to capture complex user intent or perform multi-step reasoning~\cite{shao2025reasonir}. 

\subsubsection{RAG}  
\label{sec:rag}
Retrieval-Augmented Generation (RAG)~\cite{lewis2020retrieval} integrates retrieval into the generation process by conditioning LLM responses on retrieved documents. In its standard pipeline, the model issues a query, retrieves relevant evidence, and generates an answer based on this input. While this retrieve–then–read architecture improves factual grounding, it remains limited: RAG is typically single-turn, lacks mechanisms for adaptive query refinement, and is vulnerable to irrelevant or noisy retrievals~\cite{jiang2023activerag,jin2025longcontext}. Iterative extensions~\cite{trivedi2022interleaving,asai2024self} allow multiple rounds of retrieval, but these approaches still position the LLM as a largely passive consumer of evidence rather than an active search agent.  

\subsubsection{Agentic Search}  
\label{sec:agentic_search}
Agentic search moves beyond RAG by framing the LLM as an autonomous decision-making agent. Rather than passively consuming retrieved documents, the model determines \emph{when}, \emph{where}, and \emph{how} to search, and integrates retrieved evidence into its ongoing reasoning and actions. This paradigm, often instantiated as \emph{deep research agents}~\cite{xu2025survey}, represents a shift from retrieval as static evidence injection to retrieval as dynamic tool use for problem solving. Formally, deep research agents are LLM-powered systems that integrate dynamic reasoning, adaptive planning, multi-turn data retrieval, tool use, and evidence synthesis to support complex informational research tasks.

\subsection{Basics of Reinforcement Learning}
\label{sec:rl_basic}
\begin{wrapfigure}{r}{0.28\textwidth}
  \centering
  \vspace{-1.5em} 
  \includegraphics[width=0.28\textwidth]{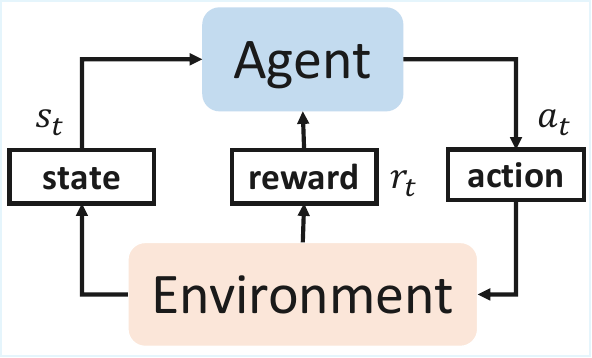}
  \vskip -1em
  \caption{Overview of RL components}
  \label{fig:rl_framework}
  \vspace{-1.5em}
\end{wrapfigure}
Reinforcement Learning (RL) is a fundamental paradigm in machine learning that studies how an agent interacts with its environment to maximize cumulative rewards through trial and error~\cite{sutton1998reinforcement}. As illustrated in Figure~\ref{fig:rl_framework}, the agent observes a state $s_t$ from the environment at each time step $t$, selects an action $a_t$ according to a policy $\pi(a_t|s_t)$, and then receives a reward $r_t$ as the environment transitions to a new state $s_{t+1}$. The agent continuously updates its policy $\pi$ to maximize the cumulative reward over time. Formally, such an optimization problem is modeled as a Markov Decision Process (MDP), represented by a tuple $(\mathcal{S}, \mathcal{A}, \mathcal{T}, \mathcal{R})$, where $\mathcal{S}$ is the set of possible states, $\mathcal{A}$ is the action space, $\mathcal{T}:\mathcal{S}\times\mathcal{A}\times\mathcal{S}\rightarrow [0,1]$ denotes the state transition probability function, and $\mathcal{R}:\mathcal{S}\times\mathcal{A}\times\mathcal{S}\rightarrow \mathbb{R}$ defines the reward function. The optimization objective is to learn a policy $\pi$ that maximizes the expected discounted cumulative reward $\sum_{k=0}^{\infty}\gamma^k r_{t+k+1}$, where $\gamma \in (0,1]$ is the discount factor.  

Policy gradient methods~\cite{schulman2017proximal,liu2025understanding,feng2025group} are widely used in RL-based agentic search, as they directly optimize stochastic policies over large discrete action spaces. Generally, they can be grouped into (i) \emph{on-policy optimization}, which updates the policy from fresh rollouts (e.g., PPO~\cite{schulman2017proximal} and GRPO~\cite{shao2024deepseekmath}); and (ii) \emph{off-policy or preference-based optimization}, which leverages offline trajectories or preference data without requiring online sampling (e.g., DPO~\cite{rafailov2023direct} and ReMix~\cite{liang2025squeeze}). 
\subsubsection{On-policy Optimization}
On-policy algorithms interact with the environment using the current policy to collect rollouts, estimate advantages, and update the same policy that generated those samples. They are favored in large-scale LLM and agentic search training due to their ability to directly optimize behavioral policies under accurate reward signals. Within this family, two subgroups can be distinguished: 
\begin{itemize}[leftmargin=*]
    \item \textbf{Critic-based algorithms.} These methods rely on an explicit \emph{value function} or \emph{critic} model to estimate the expected return for each state or token.  
    The critic provides token-level feedback that reduces the variance of policy gradients and stabilizes training, but it also introduces additional computational cost and memory overhead.
    PPO~\cite{schulman2017proximal} is the most widely used example of this paradigm.
    \item \textbf{Critic-free algorithms.} In contrast, critic-free approaches remove the value network entirely and estimate the advantage directly from relative reward statistics.  
    Instead of relying on learned value predictions, these algorithms sample multiple responses for each input and compute a \emph{group-based advantage} by normalizing rewards within the group.  
    This strategy significantly reduces training complexity and GPU memory consumption while maintaining stable optimization. Representative examples include GRPO~\cite{shao2024deepseekmath}, {Dr.GRPO}~\cite{liu2025understanding}, DAPO~\cite{yu2025dapo}, and {GiGPO}~\cite{feng2025group}.
\end{itemize}

\noindent\textbf{Proximal Policy Optimization (PPO)}. PPO~\cite{schulman2017proximal} is one of the most widely used methods for training RL agents. It aims to maximize the following objective function:
\begin{equation}\label{eq:ppo}
\small
    \mathcal{J}_{{PPO}}(\theta) = 
    \mathbb{E}_{x \sim \mathcal{D}, y \sim \pi_{{old}}( \cdot| x)} \left[
\min \left( \frac{\pi_{\theta}(y | x)}{\pi_{{old}}(y | x)} A,\right.\right.
 \left.\left. {clip}_{\epsilon} \left( \frac{\pi_{\theta}(y | x)}{\pi_{{old}}(y | x)} \right) A 
\right) -  \beta\mathbb{D}_{KL}(\pi_\theta||\pi_{ref})\right],
\end{equation}
where $\pi_{\theta}$ and $\pi_{old}$ denote the current and previous policy models, respectively. $\pi_{ref}$ is the reference model that regularizes the policy update via a KL-divergence penalty, measured and weighted by $\mathbb{D}_{KL}$ and $\beta$, respectively. $x$ denotes the input samples drawn from the distribution $D$.
$clip_{\epsilon}$ is the clipping function with hyperparameter $\epsilon$ for stabilizing training. The advantage estimate $A$ is computed using Generalized Advantage Estimation (GAE)~\citep{schulman2015high}, based on the reward $r$ and a learned value function $V_{\psi}$.

\noindent\textbf{Group Relative Policy Optimization (GRPO)}. 
GRPO~\cite{shao2024deepseekmath} extends PPO by eliminating the need for a separate value function model, which often doubles memory usage. Instead, it estimates relative advantages within groups of sampled responses from the same input, leading to improved training efficiency. Specifically, for each input $x\in D$, GRPO samples a group of outputs $\{y_1,y_2,\cdots,y_{G}\}$ from the old policy $\pi_{old}$ and optimizes the new policy $\pi_{\theta}$ by maximizing:
\begin{equation}\label{eq:grpo}
\small
    \mathcal{J}_{{GRPO}}(\theta) = 
    \mathbb{E}_{x \sim \mathcal{D}, \{y_i\}_{i=1}^{G} \sim \pi_{{old}}( \cdot| x)} 
    \frac{1}{G}\sum_{i=1}^{G} \left[
\min \left( \frac{\pi_{\theta}(y | x)}{\pi_{{old}}(y | x)} A_{i}, {clip}_{\epsilon} \left( \frac{\pi_{\theta}(y | x)}{\pi_{{old}}(y | x)} \right) A_{i} 
\right) - \beta\mathbb{D}_{KL}(\pi_\theta||\pi_{ref})\right],
\end{equation}
where $A_i$ is the advantage computed using rewards $\{r_1, r_2, \ldots, r_G\}$ corresponding to the outputs within each group:
\begin{equation}
\small
\label{eq:group_advantage}
\begin{aligned}
    A_i=\frac{r_i-\text{mean}(\{r_1,r_2,\ldots,r_{G}\})}{\text{std}(\{r_1,r_2,\ldots,r_{G}\})}.
\end{aligned}
\end{equation}

\noindent\textbf{Decoupled Clip and Dynamic Sampling Policy Optimization (DAPO)}. 
DAPO~\cite{yu2025dapo} is an emerging RL approach for training long chain-of-thought (CoT) reasoning models. Specifically, DAPO addresses several limitations of GRPO, including entropy collapse, reward noise, and training instability. It introduces four key techniques to improve RL performance in long CoT scenarios: clip-higher, dynamic sampling, token-level policy gradient loss, and overlong reward shaping. Formally, the objective function for DAPO aims to maximize the following:
\begin{equation}\label{eq:dapo}
\small
\begin{aligned}
    \mathcal{J}_{{DAPO}}(\theta) =& 
    \mathbb{E}_{x \sim \mathcal{D}, \{y_i\}_{i=1}^{G} \sim \pi_{{old}}( \cdot| x)} 
     \frac{1}{G}\sum_{i=1}^{G}\left[
\min \left( \frac{\pi_{\theta}(y | x)}{\pi_{{old}}(y | x)} A_{i}, {clip} \left( \frac{\pi_{\theta}(y | x)}{\pi_{{old}}(y | x)},  1-\epsilon_{low}, 1+\epsilon_{high} \right) A_{i} 
\right)\right] \\
& s.t., ~ 0 < |\{y_i\ |\ {it\_equivalent}(x,y_i)\}|< G,
\end{aligned}
\end{equation}
where $A_{i}$ is the advantage estimate defined in Eq.~(\ref{eq:group_advantage}). $\epsilon_{{high}}$ is typically larger than $\epsilon_{{low}}$ to provide more flexibility for increasing low-probability tokens, and ${it\_equivalent}$ is the dynamic sampling function that over-samples and filters out prompts with accuracy equal to 1 or 0. Note that the KL term is excluded in DAPO because the model distribution can diverge significantly from the initial model during the training of long CoT models.

\subsubsection{Off-policy Optimization}
Off-policy and preference-based algorithms, in contrast, do not require new rollouts from the current policy. Instead, they learn from previously collected trajectories or explicit preference annotations, which greatly improves data efficiency and stability. These methods are particularly useful in large-scale LLM alignment and agentic search scenarios, where collecting online feedback is costly or impractical.  

\noindent\textbf{Direct Preference Optimization (DPO)}. 
DPO~\cite{rafailov2023direct} is a representative \emph{RL-free} approach for aligning LLMs with human preferences. 
Unlike conventional Reinforcement Learning from Human Feedback (RLHF)~\cite{christiano2017deepreinforcementlearning,stiennon2020learningtosummarize,ouyang2022training,zhang2025bradley}, which trains a separate reward model and performs iterative policy optimization (e.g., via PPO), DPO formulates alignment as a direct probabilistic classification problem. 
It bypasses the explicit reward modeling and RL loop by learning directly from preference-labeled response pairs.
Formally, Given a dataset $\mathcal{D}$ containing triplets $(x, y_w, y_l)$, where $x$ is a prompt, and $y_w$ and $y_l$ denote the \emph{preferred (winning)} and \emph{dispreferred (losing)} responses respectively, the preferences are assumed to be generated by an underlying latent reward function $r^{*}(y, x)$ such that $r^{*}(y_w, x) > r^{*}(y_l, x)$. 
DPO optimizes the policy $\pi_{\theta}$ to increase the relative likelihood of $y_w$ over $y_l$ with respect to a reference model $\pi_{\text{ref}}$ as: 
\begin{equation}\label{eq:dpo}
\small
    \mathcal{J}_{{DPO}}(\theta) = 
    \mathbb{E}_{(x,y_w,y_l) \sim \mathcal{D}} 
    \left[
\log{\sigma} \left( \beta\frac{\pi_{\theta}(y_w | x)}{\pi_{{ref}}(y_w | x)} -  \beta\frac{\pi_{\theta}(y_l | x)}{\pi_{{ref}}(y_l | x)}\right)\right], 
\end{equation}
where $\pi_{ref}$ is the reference model, and $\beta$ is a hyperparameter that controls the strength of this regularization. The $\sigma$ function is the sigmoid, which helps to optimize the relative probability of the two responses. By using this objective, DPO directly optimizes the policy to reflect human preferences without needing an intermediate reward model.  

\begin{figure}[t]
    \centering
    \includegraphics[width=0.6\linewidth]{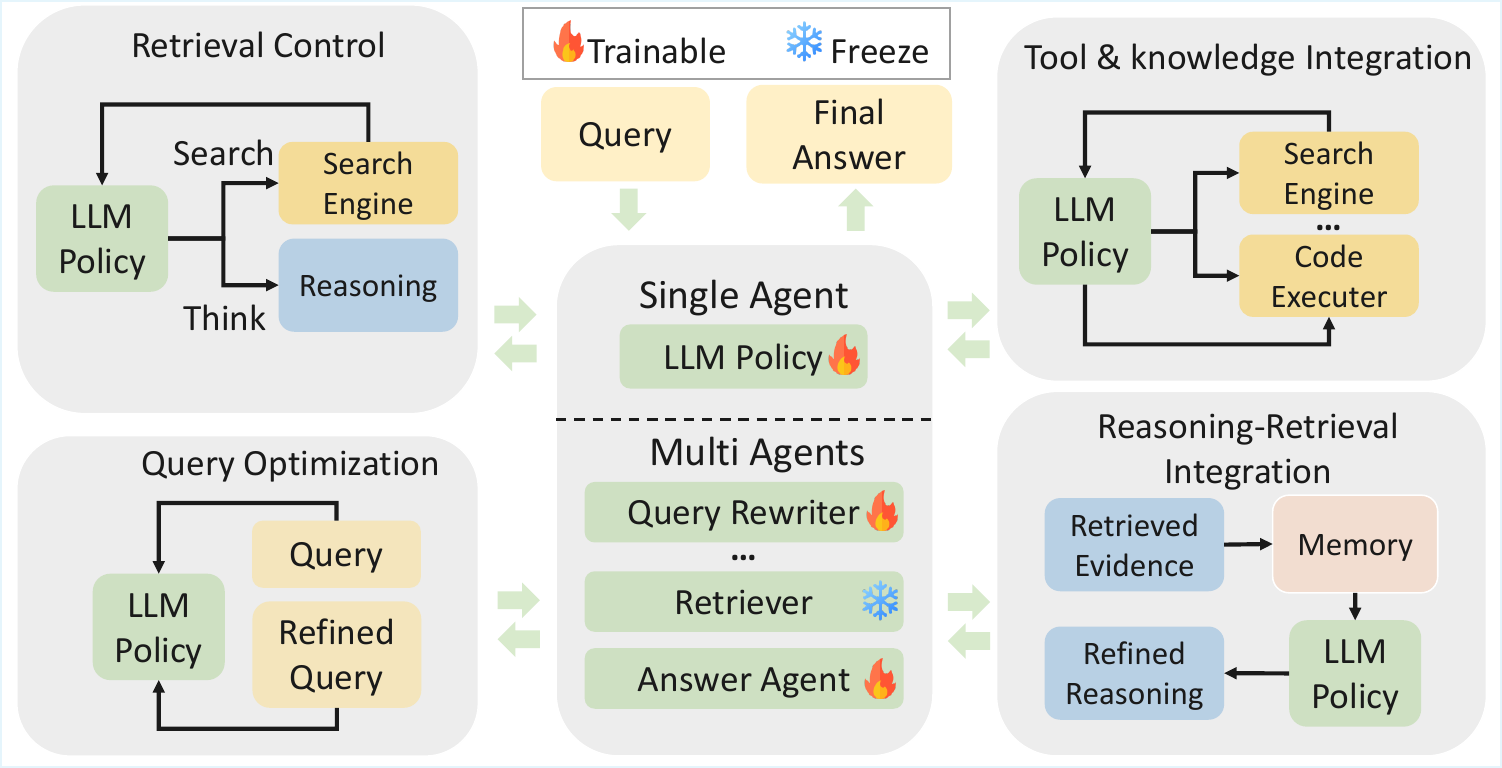}
    \vskip -0.75em
    \caption{Illustrative framework of RL-based agentic search.
    RL intervenes at multiple decision points—%
    controlling when to retrieve (retrieval control), how to formulate queries (query optimization), %
    how to integrate evidence into reasoning (reasoning-retrieval integration), %
    and which tools or knowledge sources to use (tool and knowledge integration).}
    \label{fig:rl_agentic_search_overview}
\end{figure}

\subsection{RL-based Agentic Search}
\label{sec:rl-based_agentic_search}
In agentic search, retrieval and reasoning are embedded in a \emph{sequential decision process} rather than executed as fixed, one-shot steps. The agent must decide \emph{when} to search, \emph{how} to formulate or refine queries, and \emph{how} to incorporate retrieved evidence into multi-step reasoning. Figure~\ref{fig:rl_agentic_search_overview} sketches this pipeline and highlights the decision points where RL can intervene: (i) \textbf{search control} (whether/when to retrieve), (ii) \textbf{query optimization} (how to retrieve), and (iii) \textbf{reasoning integration} (how to use retrieved information).

\subsubsection{Comparison with Pre-RL Agentic Search}
Before the introduction of RL into agentic search, most systems relied on either 
\emph{structured prompting}~\cite{zhou2024metacognitive,xu2024search,wang2023knowledge,lee2024planrag,chen2024mindsearch} 
or \emph{supervised fine-tuning (SFT)}~\cite{schick2023toolformer,asai2023self,aksitov2023rest,zhao2025tura} 
to guide retrieval and reasoning behaviors.


\noindent\textbf{Prompting-based Methods.}
These methods primarily depend on human-designed heuristics and pre-defined reasoning workflows. 
For instance, {PlanRAG}~\cite{lee2024planrag} and {MetaRAG}~\cite{zhou2024metacognitive} employ an iterative loop in which the agent alternates between searching, generating an answer, and reflecting on its quality before deciding whether to conduct further searches. This process repeats until a satisfactory response is achieved. 
Similarly, {Knowledge-driven CoT}~\cite{wang2023knowledge} follows a reflection chain that encourages the model to re-evaluate intermediate reasoning and adjust its strategy dynamically based on retrieved evidence. 
While effective, these prompting-based systems rely on fixed symbolic templates or handcrafted prompt structures that cannot adapt to unseen task distributions or dynamic retrieval environments.

\noindent\textbf{SFT-based Methods.} 
These methods train models on datasets of high-quality trajectories that include search, reflection, and generation actions, allowing the model to internalize these behaviors into its parameters. For example, Toolformer~\cite{schick2023toolformer} fine-tunes an LM on self-labeled data where API calls are automatically inserted into text generation. It learns to decide when and how to use external tools such as calculators or Wikipedia search engines, improving factuality without additional human supervision. Similarly, SelfRAG~\cite{asai2023self} introduces \emph{self-reflective retrieval-augmented generation}, where the model is supervised to generate both normal tokens and special \emph{reflection tokens} (e.g., \texttt{<Retrieve>}, \texttt{<Relevant>}, \texttt{<Supported>}) that indicate when to retrieve new evidence and how well each generation is supported by retrieved passages. 
Despite these advances, SFT-based approaches remain fundamentally imitation-driven. They can capture correlations between context and actions but lack mechanisms for long-horizon credit assignment or outcome-driven optimization.

\noindent\textbf{Limitations and Why RL.}
Despite their progress, both prompting- and SFT-based agents face inherent limitations:
\begin{itemize}[leftmargin=*]
    \item \emph{Poor adaptivity}: Their behaviors are largely predefined or imitated from static datasets. They cannot dynamically adjust retrieval frequency or reformulate queries when facing unseen tasks or API behaviors.
    \item \emph{Supervision bottleneck}: High-quality reasoning and search trajectories are costly to collect and difficult to scale across tasks, which constrains generalization and makes further improvement beyond demonstrations challenging.
\end{itemize}
RL provides a principled way to overcome these issues by optimizing the agent as a policy $\pi_\theta$ that interacts with an environment, receives feedback, and adapts through trial and error.  
Unlike SFT-based imitation, RL directly optimizes task-level rewards that integrate correctness, cost, and latency, enabling the discovery of \emph{adaptive and efficient} retrieval policies.  
This paradigm allows the agent to reason about the \emph{long-term consequences} of each search decision, moving beyond static imitation toward outcome-driven learning.

\subsubsection{Formalization.} 
Formally, RL-based agentic search can be modeled as a MDP. 
The goal is to train a policy $\pi_\theta$ that maximizes cumulative reward by taking a sequence of actions in an environment. 
The key components are: (i) \textbf{Agent}: The LLM policy $\pi_\theta$, parameterized by $\theta$, which generates actions conditioned on the current state; (ii) \textbf{Environment}: External resources the agent can interact with, such as search engine APIs, retrievers, knowledge graphs, or tool interfaces; (iii) \textbf{State ($s_t$)}: The current context, including the original query, intermediate reasoning traces, retrieved evidence, and action history; (iv) \textbf{Action ($a_t$)}: A discrete decision, such as issuing a query, reformulating an existing query, selecting documents, invoking tools (e.g., search APIs, retrievers), or terminating with a final answer; (v) \textbf{Action ($a_t$)}: A discrete decision, such as issuing a query, reformulating an existing query, selecting documents, invoking tools (e.g., search APIs, retrievers), or terminating with a final answer; (vi) \textbf{Reward ($r_t$)}: A scalar feedback signal capturing task success (e.g., answer correctness, factual consistency), process quality (e.g., query efficiency, reasoning coherence), or resource costs (e.g., API calls, latency); and (vii) \textbf{Transition ($\mathcal{T}$)}: The dynamics induced by both the environment (e.g., a search engine returning documents) and the agent’s internal updates.

%% file: 5_What_is_for_v3.tex
\begin{table}[!t]
\centering
\small
\setlength{\tabcolsep}{4pt}
\renewcommand{\arraystretch}{1.15}
\caption{The categorization of RL-based search agents from functional roles' perspective.}
\label{tab:what_is_for}
\vskip -1em
\begin{tabularx}{\textwidth}{
>{\RaggedRight\arraybackslash}p{2.5cm}
>{\RaggedRight\arraybackslash}p{4.1cm}
>{\RaggedRight\arraybackslash}X
}

\toprule
\textbf{Category} & \textbf{Functional Roles} & \textbf{Methods} \\
\midrule

\multirow{9}{*}{\textbf{Retrieval Control}}
&\multirow{3}{*}{Adaptive search decisions}
&
Search-R1~\cite{jin2025searchr1}; ReSearch~\cite{chen2025learning}; DeepRAG~\cite{guan2025deeprag}; UR$^2$~\cite{li2025ur}; SSRL~\cite{fan2025ssrl}; R1-Searcher~\cite{song2025r1}; AutoCoA~\cite{zhang2025agent};DeepNote~\cite{wang2024retriever}; SWiRL~\cite{goldie2025synthetic};DeepResearcher~\cite{zheng2025deepresearcher}; MedResearcher-R1~\cite{yu2025medresearcher} \\
\cmidrule{2-3}
&
\multirow{2}{*}{Search intensity and persistence}
&
Pangu DeepDiver~\cite{shi2025pangu}; ReZero~\cite{dao2025rezero}; StepSearch~\cite{wang2025stepsearch}; ReasonRAG~\cite{zhang2025process}; WebSailor-V2~\cite{li2025websailorv2} \\
\cmidrule{2-3}
&
\multirow{4}{*}{Search efficiency}
&
IKEA~\cite{huang2025reinforced}; R1-Searcher++~\cite{song2025r1++}; 
DeepRAG~\cite{guan2025deeprag}; 
Search Wisely~\cite{wu2025search}; StepSearch~\cite{wang2025stepsearch}; ZeroSearch~\cite{sun2025zerosearch}; ParallelSearch~\cite{zhao2025parallelsearch}; RAG-R1~\cite{tan2025ragr1}; ReasonRAG~\cite{zhang2025process};
WebThinker~\cite{li2025webthinker};DeepResearcher~\cite{zheng2025deepresearcher} \\
\midrule

\multirow{4}{*}{\textbf{Query Optimization}}
&
\multirow{2}{*}{Conversational reformulation}
&
ConvSearch-R1~\cite{zhu2025convsearch}; MaskSearch~\cite{wu2025masksearch}; RAG-R1~\cite{tan2025ragr1}; ParallelSearch~\cite{zhao2025parallelsearch}; OPERA~\cite{liu2025opera}; WebExplorer~\cite{liu2025webexplorer}; DeepNote~\cite{wang2024retriever} \\
\cmidrule{2-3}
&
\multirow{2}{*}{Retriever-aware optimization}
&
DeepRetrieval~\cite{jiang2025deepretrieval}; ZeroSearch~\cite{sun2025zerosearch}; s3~\cite{jiang2025s3}; WebThinker~\cite{li2025webthinker}; MMOA-RAG~\cite{chen2025improving} \\
\midrule

\multirow{4}{*}{\makecell[l]{\textbf{Reasoning–Retrieval}\\\textbf{Integration}}}
&
Reasoning–search interleaving 
&
SWiRL~\cite{goldie2025synthetic}
R-Search~\cite{zhao2025r}; AutoRefine~\cite{shi2025search}; EvolveSearch~\cite{zhang2025evolvesearch}; ReasonRAG~\cite{zhang2025process}; O${^2}$-Searcher~\cite{mei2025o2};Atom-Searcher~\cite{deng2025atom} \\
\cmidrule{2-3}
&
Context and memory management
&
ReSum~\cite{wu2025resum}; SFR-DeepResearch~\cite{nguyen2025sfr};
DeepResearcher~\cite{zheng2025deepresearcher}; RECON~\cite{xu2025reconreasoningcondensationefficient}; WebSailor~\cite{li2025websailor}; WebSailor-V2~\cite{li2025websailorv2}; ASearcher~\cite{gao2025beyond}
\\
\midrule

\multirow{3}{*}{\makecell[l]{\textbf{Multi-Agent}\\\textbf{Collaboration}}}
&
\multirow{1}{*}{Planner–executor orchestration}
&
MAO-ARAG~\cite{chen2025mao}; OPERA~\cite{liu2025opera}; AI-SearchPlanner~\cite{mei2025ai} \\
\cmidrule{2-3}
&
\multirow{2}{*}{Cooperative multi-agent systems}
&
SIRAG~\cite{wang2025sirag}; MMOA-RAG~\cite{chen2025improving}; AgentGym-RL~\cite{xi2025agentgym}; Chain-of-Agents~\cite{li2025chain}; WebExplorer~\cite{liu2025webexplorer} \\
\midrule

\multirow{6}{*}{\makecell[l]{\textbf{Tool and Knowledge}\\\textbf{Integration}}}
&
\multirow{3}{*}{Multi-tool}
&
Tool-Star~\cite{dong2025tool}; VerlTool~\cite{jiang2025verltool}; WebWatcher~\cite{geng2025webwatcher}; AI-SearchPlanner~\cite{mei2025ai}; WebSailor-V2~\cite{li2025websailorv2}; WebResearcher~\cite{qiao2025webresearcher}; MedResearcher-R1~\cite{yu2025medresearcher}\\
\cmidrule{2-3}
&
\multirow{2}{*}{Multi-modality}
&
Visual-ARFT~\cite{liu2025visual}; VRAG-RL~\cite{wang2025vrag}; MMSearch-R1~\cite{wu2025mmsearch}; WebWatcher~\cite{geng2025webwatcher}\\
\cmidrule{2-3}
&
\multirow{1}{*}{Structured knowledge navigation}
&
GRAIL~\cite{chang2025grail}; DynaSearcher~\cite{hao2025dynasearcher} \\
\bottomrule
\end{tabularx}

\end{table}

\section{What RL is for: Functional Roles in Agentic Search}
\label{sec:what_RL_is_for}
RL plays a wide range of functional roles within agentic search, extending well beyond basic retrieval. In this section, we categorize these roles into five major dimensions to illustrate how RL enables agents to decide not only \emph{when} to search, but also \emph{how} to formulate queries, \emph{how} to interleave reasoning with evidence, and \emph{how} to coordinate across multiple agents and tools. Table~\ref{tab:what_is_for} summarizes representative works of each RL's role.


\subsection{Retrieval Control}
\label{sec:retrieval_control}

A core role of RL in agentic search is to control \emph{whether, when, and how} an agent retrieves external information.  
Rather than being a fixed design principle, this perspective synthesizes recent trends observed across RL-based retrieval systems~\cite{huang2025reinforced,jin2025searchr1,wu2025search,wang2025stepsearch}, where retrieval control emerges as a central optimization target.  
Effective retrieval control is crucial, since excessive or unnecessary queries increase cost and latency, while insufficient retrieval risks missing critical evidence.  
RL enables agents to balance this trade-off by learning adaptive retrieval policies that respond to task context and uncertainty.  
Methods in this category address three key aspects: (i) \emph{adaptive search decisions}—whether to retrieve or rely on parametric knowledge, (ii) \emph{search intensity and persistence}—how often and how deeply to retrieve, and (iii) \emph{search efficiency}—minimizing redundancy, cost, and latency while preserving task performance.


\subsubsection{Adaptive Search Decisions}
\label{sec:adaptive search decision}
RL enables agents to decide whether a question can be answered using internal parametric knowledge or requires external retrieval.
Search-R1~\cite{jin2025searchr1}, ReSearch~\cite{chen2025learning}, and R1-Searcher~\cite{song2025r1} are early examples that teach LLMs to invoke search engines only when necessary.
Specifically, as shown in Table~\ref{tab:search-r1_prompt}, these methods encourage LLMs to call a search engine to access external information when the internal knowledge is insufficient to produce an accurate answer.
Building on this idea, DeepRAG~\cite{guan2025deeprag} formulates RAG as a MDP, where complex queries are \emph{iteratively decomposed into atomic subqueries}, each representing a focused information need.
At each reasoning step, the agent decides whether to answer the subquery using its parametric knowledge or to retrieve external evidence, guided by a reward that jointly optimizes answer correctness and retrieval cost.

\subsubsection{Search Intensity}
\label{sec:search_intensity}
For complex or ambiguous queries, a single retrieval attempt may be insufficient. RL has been used to optimize the depth and persistence of the search process.  
Pangu DeepDiver~\cite{shi2025pangu} introduces \emph{Search Intensity Scaling}, rewarding agents for intensifying retrieval when ambiguity is detected.  
ReZero~\cite{dao2025rezero} rewards retry attempts after failed searches, encouraging persistence and robustness.  
StepSearch~\cite{wang2025stepsearch} introduces step-wise rewards based on information gain and redundancy penalties to guide retrieval step by step. 


\subsubsection{Search Efficiency.} 
\label{sec:search_efficiency}
Efficiency concerns both the \emph{cost} of retrieval (e.g., number of API calls, training rollouts) and the \emph{time} required to complete searches. 
R1-Searcher++~\cite{song2025r1++} extends R1-Searcher by introducing a \emph{group reward} that measures retrieval thriftiness through the variance of retrieval counts across responses, rewarding the correct answer that requires the fewest retrieval calls while penalizing redundant searches. 
IKEA~\cite{huang2025reinforced} introduces knowledge-boundary–aware rewards that favor internal reasoning unless external retrieval is necessary. 
Search Wisely~\cite{wu2025search} improves cost efficiency by filtering low-confidence queries that are likely to yield poor results. 
StepSearch~\cite{wang2025stepsearch} penalizes redundant queries with step-wise rewards, encouraging more concise retrieval strategies. 
ZeroSearch~\cite{sun2025zerosearch} reduces API overhead by simulating retrieval in latent space, enabling curriculum-style training without reliance on real search engines. 
Beyond reducing retrieval calls, ParallelSearch~\cite{zhao2025parallelsearch} decomposes complex questions into parallel sub-queries to maintain coverage while significantly lowering response time, and RAG-R1~\cite{tan2025ragr1} similarly incentivizes multi-query parallelism to enhance inference efficiency.
In addition, WebThinker~\cite{li2025webthinker} extends the notion of efficiency from search cost to reasoning behavior, applying preference optimization to align query strategies with long-horizon reasoning objectives such as correctness, tool efficiency, and thinking conciseness, thereby refining retrieval decisions through reasoning-driven feedback rather than retrieval accuracy alone.


\subsection{Query Optimization}
\label{sec:query_optimization}
Even when retrieval is triggered, the quality of queries strongly influences outcomes. Poorly posed queries yield irrelevant or noisy results. RL is then used to refine query generation based on feedback, moving beyond static heuristics.  Existing works can be categorized into  (i) \emph{conversational reformulation} and (ii) \emph{retriever-aware optimization}.


\subsubsection{Conversational Reformulation}  
\label{sec:conver_reformulation}
In interactive settings, user queries are often ambiguous or context-dependent, making direct retrieval unreliable. RL enables agents to reformulate such inputs into self-contained queries by framing reformulation as a sequential decision-making process. ConvSearch-R1~\cite{zhu2025convsearch} optimizes a rewriter policy with retrieval-based rewards, where higher rewards are assigned when reformulated queries retrieve gold passages at higher ranks. Its rewriter is first fine-tuned through SFT on data generated via retrieval-guided self-distillation, and then refined through RL using a \emph{Rank-Incentive Reward Shaping} function that encourages ranking gold passages higher while mitigating reward sparsity. This two-stage design aligns the query rewriter with retriever preferences and improves retrieval precision in multi-turn search. MaskSearch~\cite{wu2025masksearch} 
extends this paradigm by incorporating a \emph{Rewriter Agent} to refine search queries for more comprehensive retrieval, whose outputs are further used in the reasoning traces for the SFT of the LLM. Instead of optimizing a separate rewriter policy, RAG-R1~\cite{tan2025ragr1} encourages the LLM itself to generate multiple parallel queries within a single prompt to improve inference efficiency and retrieval diversity. Similarly, ParallelSearch~\cite{zhao2025parallelsearch} trains LLMs to decompose complex or multi-hop questions into parallel sub-queries within a single reasoning turn. During RL fine-tuning, a \emph{decomposition reward} encourages effective query breakdown, while a \emph{search-count reward} penalizes excessive search actions, balancing reformulation granularity and retrieval efficiency.

\subsubsection{Retriever-Aware Optimization}
\label{sec:retriever_optimization}
While conversational reformulation focuses on resolving user-side ambiguity, retriever-aware optimization instead targets the system side of query generation. It trains agents to adapt their queries to the characteristics, biases, and feedback signals of specific retrievers. The objective is to bridge the semantic gap between LLM-generated queries and the retriever’s actual ranking behavior, thereby improving retrieval accuracy and robustness across different search infrastructures.  
DeepRetrieval~\cite{jiang2025deepretrieval} exemplifies this idea by training LLMs to produce queries that align with the biases of black-box search engines, effectively exploiting retriever behavior to maximize recall.   
WebThinker~\cite{li2025webthinker} applies preference optimization to align query strategies with long-horizon reasoning objectives such as correctness, tool efficiency, and thinking conciseness, enabling the agent to refine its search behavior using reasoning-driven feedback instead of retrieval accuracy alone.
ZeroSearch~\cite{sun2025zerosearch} further extends this approach by simulating retrieval environments, allowing agents to learn robust query behaviors that generalize across different retrievers while avoiding the cost and instability of real API calls.  
Similarly, s3~\cite{jiang2025s3} introduces a lightweight RL-based searcher module decoupled from the LLM generator, enabling scalable and model-agnostic query optimization.  
Together, these approaches highlight the broader goal of designing retriever-aware query policies that remain effective across heterogeneous search environments.

\subsection{Reasoning–Retrieval Integration}  
\label{sec:reasoning-retrieval-integration}
Beyond deciding \emph{when} and \emph{how} to search effectively, knowledge-intensive tasks often require tight coupling between reasoning and retrieval. 
Evidence is only valuable if it improves reasoning, and reasoning should guide what to retrieve next. RL optimizes how LLMs interleave these processes, manage context, and refine reasoning based on feedback.  


\subsubsection{Reasoning–Search Interleaving} 
\label{sec:reasoning_search_interleaving}
Beyond simply allowing retrieval during reasoning~\cite{jin2025searchr1,chen2025learning}, RL optimizes retrieval to enhance reasoning quality.  
R-Search~\cite{zhao2025r} introduces an \emph{evidence reward} to encourage high-quality query generation yielding more informative evidences.  
AutoRefine~\cite{shi2025search} extends the standard ``search-and-think'' paradigm to ``search-and-refine-during-think,'' rewarding intermediate refinement steps to reinforce faithful and targeted knowledge extraction. 
EvolveSearch~\cite{zhang2025evolvesearch} further strengthens reasoning–retrieval interplay through iterative cycles of SFT and RL to enhance the data efficiency during training, enabling agents to progressively refine both their reasoning paths and retrieval strategies.  
In contrast, MaskSearch~\cite{wu2025masksearch} focuses on enhancing the model’s retrieval-aware reasoning ability \emph{before} RL optimization. It introduces a \emph{Retrieval-Augmented Mask Prediction (RAMP) pretraining task}, which teaches the model to leverage external search tools to fill masked spans with retrieved knowledge in the SFT stage. This pre-RL objective establishes a retrieval-aware prior that aligns reasoning and retrieval behaviors, enhancing the universal search capabilities across various downstream tasks.

\subsubsection{Context and Memory Management}
\label{sec:context_memory_management}
While existing agentic search systems~\cite{zhao2025r,jin2025searchr1,wang2025stepsearch} are effective for short-horizon tasks such as single-turn retrieval or step-level reasoning, they often struggle in long-horizon or multi-session settings, where agents must manage extended interaction histories within limited context windows. 
To operate efficiently under such constraints, agents need to \emph{actively manage memory}—deciding what information to retain, summarize, or discard as a search episode unfolds. 
Recent studies~\cite{gao2025beyond,xu2025reconreasoningcondensationefficient,wu2025resum,chen2025sft,li2025websailor,li2025websailorv2} apply RL to optimize this process, framing memory control as a sequential decision problem balancing \emph{information fidelity} and \emph{context efficiency}. Specifically, two complementary strategies have emerged:
\begin{itemize}[leftmargin=*]
    \item \textbf{Internal management:} The agent itself performs memory operations such as summarizing, refreshing, or pruning its working context under RL guidance. 
    For instance, {ReSum}~\cite{wu2025resum} trains agents with RL to generate concise summaries of past reasoning and interactions, enabling long-context reasoning without exceeding token limits. 
    {SFR-DeepResearch}~\cite{nguyen2025sfr} further introduces explicit memory actions (e.g., \texttt{clean\_memory}, \texttt{store\_snippet}), using RL signals to decide when to retain or discard past information, thus preventing memory overflow and redundancy.

    \item \textbf{External management:} Other frameworks use auxiliary summarization modules to compress historical context before reinjection into the agent’s reasoning stream. 
    In such cases, RL or policy learning is used to determine when and how to invoke these summarizers. 
    For example, \textbf{WebSailor}~\cite{li2025websailor} employs an external summarizer to condense browsing traces for multi-page search; 
    \textbf{ASearcher}~\cite{gao2025beyond} dynamically summarizes multi-turn research sessions to preserve key findings; and     \textbf{RECON}~\cite{xu2025reconreasoningcondensationefficient} integrates a frozen, pretrained summarizer into an RL-based search agent (e.g., Search-R1); the summarizer, trained via supervised relevance pretraining and multi-aspect distillation, enables the agent to reason over concise, factual evidence while substantially reducing context length and cost.
\end{itemize}

\subsection{Multi-Agent Collaboration}
\label{sec:what_is_for_multi_agent_collaboration}
Beyond relying on a single LLM to handle both reasoning and retrieval, advanced agentic search systems~\cite{chen2025mao, wang2025sirag} decompose the process into multiple specialized modules, such as query rewriting~\cite{ma2023query}, document selection~\cite{ke2024bridging}, and reasoning control.  
RL is then used to align the objectives of distinct agents, ensuring that local decisions, such as when to reformulate, which evidence to retain, and how to schedule retrieval steps, contribute to globally coherent and efficient search.  
Existing approaches can be broadly categorized into (i) \emph{planner–executor architectures} and (ii) \emph{cooperative multi-agent systems}.

\begin{figure}[t]
    \centering
    \includegraphics[width=1\linewidth]{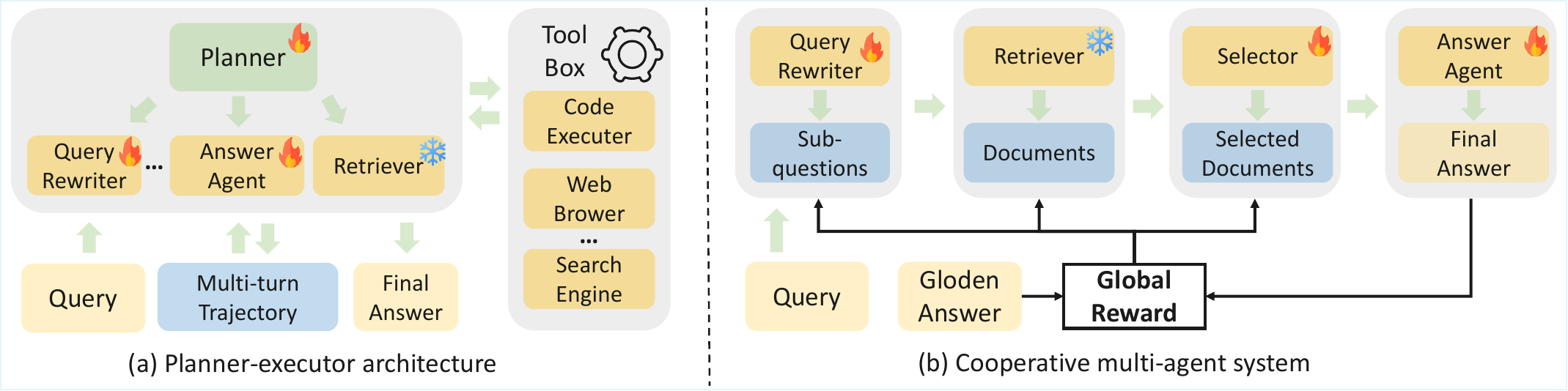}
    \vskip -1em
    \caption{Overview of RL for multi-agent collaboration. (a)~\emph{Planner--executor architecture}: a central planner coordinates specialized executor agents
    for task decomposition and dynamic subtask allocation.
    (b)~\emph{Cooperative multi-agent system}: multiple agents jointly optimize shared objectives
    through communication, coordination, and reward sharing.}
    \label{fig:multi_agent_search_framework}
\end{figure}

\subsubsection{Planner–Executor Architectures} 
\label{sec:planner_executor_architectures}
A representative paradigm is the \emph{planner–executor architecture}, where a high-level planner orchestrates specialized executors responsible for distinct retrieval or reasoning operations. As shown in Figure~\ref{fig:multi_agent_search_framework}(a),  the planner acts as a meta-policy that decides which executor to invoke, when to switch subtasks, and how to allocate search or computational budgets, thus achieving \emph{adaptive orchestration} across heterogeneous RAG modules.

\textbf{MAO-ARAG}~\cite{chen2025mao} exemplifies this design. It models multi-agent RAG as a \emph{multi-agent semi-Markov decision process (MSMDP)}, where the planner coordinates executors such as query rewriters, document selectors, retrievers, and generators. Specifically, a planner agent intelligently se-
lects and integrates the appropriate agents from these execu-
tors into a suitable workflow tailored for each query, striving
for high-quality answers while maintaining reasonable costs. During each turn, the planner agent is trained using PPO, optimizing by the following reward:
\begin{equation}
    r_t = r_{\text{F1}} - \alpha \cdot r_{\text{CP}}  - r_{\text{FP}},
\end{equation}
where $r_{\text{F1}}$ is the outcome-based reward based on F1 score, and $r_{\text{CP}}$ and $r_{\text{FP}}$ are the cost penalty and format penalty, respectively. These rewards together improve answer quality while keeping costs within a reasonable range.

{OPERA}~\cite{liu2025opera} extends this idea to multi-hop retrieval and reasoning.  It adopts a hierarchical RL framework composed of a high-level planning module and low-level execution agents. Three role-specific agents, including \emph{Plan}, \emph{Analysis–Answer}, and \emph{Rewrite}, are optimized with {Multi-Agents Progressive GRPO (MAPGRPO)}, a GRPO-based algorithm that provides fine-grained, role-specific credit assignment. Each agent is trained with tailored reward signals: the Plan Agent for decomposition validity, the Analysis–Answer Agent for reasoning and factual correctness, and the Rewrite Agent for retrieval relevance and formatting. 
This hierarchical optimization yields stable convergence and interpretable reasoning trajectories, enabling OPERA to learn cost-efficient, verifiable retrieval–reasoning workflows.

\subsubsection{Cooperative Multi-Agent Systems}
\label{sec:cooperative_multi-agent_system}
Another workflow models the agentic search as a cooperative muti-agent games, where each module is treated as an RL agent whose actions influence retrieval outcomes, with a shared global reward aligning their behaviors toward better performance. The overall framework is illustrated in Figure~\ref{fig:rl_agentic_search_overview}(b).
For example,
SIRAG~\cite{wang2025sirag} trains a Decision-Maker to decide when to retrieve and a Knowledge-Selector to filter which documents should be passed downstream, with RL rewards aligning their decisions toward high-quality evidence integration. 
MMOA-RAG~\cite{chen2025improving} generalizes this setting to larger agent pools, where RL optimizes how agents share responsibilities for query reformulation, evidence selection, and verification. 
In addition, some works such as AgentGym-RL~\cite{xi2025agentgym} and Chain-of-Agents~\cite{li2025chain} provide general infrastructures for training multi-agent systems, where agentic search is a core evaluation setting.


\subsection{Tool and Knowledge Integration}
\label{sec:tool_knowledge_integration}
Finally, rather than relying solely on text retrieval, agentic search increasingly requires integration with heterogeneous external resources, including APIs~\cite{jiang2025verltool}, multi-modal tools~\cite{wang2025vrag, geng2025webwatcher}, and structured knowledge bases~\cite{chang2025grail, hao2025dynasearcher}, to extend the scope of tasks agents can solve, where RL is a natural solution to enable them. 
Research in this category can be grouped into two directions: (i) \emph{multi-tool and multi-modality reasoning}, where agents learn to coordinate across diverse toolkits such as search engines, code interpreters, and vision models, and (ii) \emph{structured knowledge exploration}, where RL trains agents to navigate symbolic environments like knowledge graphs or tables in a goal-directed way.  

\subsubsection{Multi-tool and Multi-modality Reasoning}
\label{sec:multi_tool_multi_modal_reasoning}
Many tasks require more than text-based retrieval, demanding agents to combine computation, web search, and multimodal understanding. RL has been used to optimize tool selection and sequencing by providing feedback on whether tool calls lead to accurate reasoning or task completion. Tool-Star~\cite{dong2025tool} integrates six tools, including search engines and code generators, using a self-critic RL setup that rewards correct intermediate outputs. VerlTool~\cite{jiang2025verltool} generalizes this with a unified RL framework that manages heterogeneous APIs and multi-modal LLMs (MLLMs).  
In multi-modal contexts, MMSearch-R1~\cite{wu2025mmsearch}, Visual-ARFT~\cite{liu2025visual}, and VRAG-RL~\cite{wang2025vrag} extend Search-R1 paradigms to visual question answering by rewarding policies that align retrieved text and visual evidence. WebWatcher~\cite{geng2025webwatcher} further trains agents with RL to coordinate multiple tools simultaneously, handling both textual and visual inputs.  

\subsubsection{Structured Knowledge Navigation}
\label{sec:structure_knowledge_navigatino}
In many domains, critical information is stored in structured resources such as knowledge graphs (KG) or databases~\cite{bollacker2008freebase, zhong2017seq2sql,lin2024decoding,lin2025stealing}. RL is applied by defining traversal as a sequential decision-making process: each step selects which entity or relation to follow, with rewards reflecting correctness, coverage, or efficiency. For instance, GRAIL~\cite{chang2025grail} applies RL to learn KG traversal policies that reach correct answers efficiently. DynaSearcher~\cite{hao2025dynasearcher} extends this with multi-reward RL, jointly optimizing for accuracy, efficiency, and balanced exploration of KG.


\begin{myboxi}
\textbf{Takeaways}: 
\begin{itemize}[leftmargin=*]
  \item \textbf{Retrieval Control}: RL improves when to retrieve, how persistently to search, and how to minimize cost and latency. Current limitations include the reliance on narrow reward signals (often correctness-only) and evaluations confined to controlled settings, which limit robustness in real, noisy retrieval environments.
  \item \textbf{Query Optimization}: RL enables both query reformulation and retriever-aware adaptation, improving retrieval precision. A key gap is generalization beyond static datasets, simulators, or single-retriever setups.
  \item \textbf{Reasoning–Retrieval Integration}: RL also extends beyond retrieval control to jointly optimize reasoning and evidence use, while also empowering active memory management such as summarizing, refreshing, and pruning; yet most memory mechanisms remain heuristic and struggle with long-term continuity. 
  \item \textbf{Multi-Agent Collaboration}: RL aligns planner–executor and cooperative agents so local actions (reformulate, select, verify) serve global objectives, improving division of labor and consistency in complex pipelines. 
  \item \textbf{Tool \& Knowledge Integration}: RL allows agents to coordinate heterogeneous tools and structured knowledge sources beyond text-only retrieval, although current systems remain at an early stage and face challenges in maintaining coherent reasoning across modalities and asynchronous feedback.
\end{itemize}
Overall, these roles form a continuum that spans from \emph{when to retrieve} through \emph{how to query} and \emph{how to think with evidence}, to \emph{who coordinates} and \emph{which tools or knowledge bases to use}, revealing RL as the unifying mechanism that grounds, scales, and organizes agentic search behaviors.
\end{myboxi}


%% file: 4_How_to_use_v3.tex
\section{How RL is Used: Optimization Strategies}
\label{sec:how_RL_is_used}
This section examines how RL is applied in agentic search systems, covering training pipelines, algorithmic design, and reward mechanisms. Table~\ref{tab:how_to_use} summarizes representative works with corresponding optimization strategies.
\subsection{Training Regime}
\label{sec:training_regime}
The training regime defines how RL is integrated into agentic search, encompassing initialization strategies, environment design, and optimization workflows. It determines how agents acquire, refine, and stabilize their decision-making policies throughout interaction-based learning.

\subsubsection{Standard Agentic Search Pipeline} 
\label{sec:standard_agentic_search}
A typical RL training pipeline for agentic search, exemplified by Search-R1~\cite{jin2025searchr1}, comprises two stages: a \emph{cold-start} initialization and subsequent RL fine-tuning.  
The cold-start phase ensures interface compliance (e.g., API calls, tool schemas) and stabilizes early rollouts.  
During RL training, the policy LLM receives complex queries and generates interleaved reasoning and tool-use actions within simulated or real search environments.  
The overall training pipeline and prompt template are summarized in Table~\ref{tab:search-r1_prompt}.
\begin{table*}[!t]
    \centering
    \caption{Standard agentic search prompt template. We use the prompt template of Search-R1~\cite{jin2025searchr1} as an example.}
    \label{tab:search-r1_prompt}
    \vskip -1em
    \fontsize{9pt}{11pt}\selectfont
    \begin{tabular}{p{0.98\linewidth}}
    \midrule
        \rowcolor{gray!20}\textbf{Search-R1 Promp Template} \\
    \midrule
Answer the given question. You must conduct reasoning inside \texttt{<think>} and \texttt{</think>} first every time you get new information. After reasoning, if you find you lack some knowledge, you can call a search engine by \texttt{<search>} query \texttt{</search>}, and it will return the top searched results between \texttt{<information>} and \texttt{</information>}. You can search as many times as you want. If you find no further external knowledge needed, you can directly provide the answer inside \texttt{<answer>} answer \texttt{</answer>} without detailed illustrations. For example, \texttt{<think>} xxx \texttt{</think>}. Question: question.\\
\midrule
\end{tabular}
\end{table*}


\subsubsection{Cold Start}
\label{sec:cold_start}
A dominant paradigm initializes agents via supervised fine-tuning (SFT) before RL optimization~\cite{li2025websailor,wu2025webdancer,dong2025tool,song2025r1++}. This stage equips models with baseline task competence and mitigates early instability caused by sparse rewards in long-horizon environments. For instance, Webagent-R1~\cite{wei2025webagent} shows that SFT provides crucial web-interaction knowledge for downstream RL, while WebSailor~\cite{li2025websailor} finds that SFT accelerates convergence and stabilizes multi-step tool use.  
EvolveSearch~\cite{zhang2025evolvesearch} further introduces a self-improving SFT–RL loop, where RL-refined policies generate new demonstrations for iterative SFT retraining.  
Conversely, several works~\cite{sun2025zerosearch,xi2025agentgym} question the necessity of SFT. ZeroSearch~\cite{sun2025zerosearch} replaces it with latent-space retrieval simulation, enabling pure RL training without external supervision, while AgentGym-RL~\cite{xi2025agentgym} employs curriculum-based horizon scaling to stabilize RL-only training.  

\subsubsection{Simulation-Based Training} 
\label{sec:simulation_training}
Training RL agents in real-world search environments can be prohibitively expensive, slow, and non-reproducible. Simulation environments provide a controlled, accelerated, and cost-effective alternative.
For example, ZeroSearch~\cite{sun2025zerosearch} proposes a novel RL framework that \emph{simulates} search by transforming an LLM into a retrieval module, avoiding the cost and noise of real search engines during training. It employs a curriculum that incrementally degrades the quality of simulated documents, forcing the agent to become more robust. O$^2$-Searcher~\cite{mei2025o2} also leverages an efficient, locally simulated search environment for training, focusing on open-domain open-ended question answering scenarios.
WebSailor-V2~\cite{li2025websailorv2} proposes a dual-environment RL framework, utilizing a high-fidelity simulator for rapid algorithm iteration and a robust, managed real-world environment for stable final policy training. This hybrid approach addresses the challenges of both scalability and realism.


\subsubsection{RL Algorithms}
\label{sec:RL_algorithms_in_how_to_use}
Most RL-based search agents employ policy-gradient algorithms, particularly PPO~\cite{schulman2017proximal}, GRPO~\cite{shao2024deepseekmath}, and Reinforce++~\cite{hu2025reinforce++}.  
Recent variants adapt these methods to the search context: Search Wisely~\cite{wu2025search} introduces $\beta$-GRPO for uncertainty-aware calibration, StepSearch~\cite{wang2025stepsearch} implements step-wise PPO aligned with information gain, and ReinforceRAG~\cite{zeng2025reinforcing} augments policy gradients with retrieval-aware baselines to mitigate variance under sparse rewards.  
The details of the RL algorithms applied in RL-based search agents are in Table~\ref{tab:how_to_use}.

\subsubsection{Curriculum Learning and Horizon Scaling}
\label{sec:curriculum}
RL training for long-horizon search tasks remains challenging due to sparse rewards and unstable credit assignment. Curriculum learning alleviates these issues by gradually expanding task complexity or interaction length.  
AgentGym-RL~\cite{xi2025agentgym} proposes \emph{ScalingInter-RL}, which progressively extends the interaction horizon—starting from short, focused tasks and gradually scaling to multi-step reasoning—balancing exploration and exploitation.  
ZeroSearch~\cite{sun2025zerosearch} employs a curriculum that systematically increases retrieval noise, compelling agents to develop more resilient strategies.  
InfoSeek~\cite{xia2025open} similarly generates progressively harder research tasks to facilitate structured capability growth.  
These strategies jointly improve convergence stability and support continual capability scaling.


\subsubsection{Iterative and Self-Evolving Frameworks}
\label{sec:iterative_self_evolving}
Beyond static curricula, some frameworks close the loop between data generation and policy learning. EvolveSearch~\cite{zhang2025evolvesearch} epitomizes this approach: RL-trained models generate higher-quality search trajectories that are distilled back into SFT data, creating a self-reinforcing cycle of improvement.  
Such iterative frameworks demonstrate how RL can act not only as a training objective but as a data generator, continuously refining both model behavior and supervision quality.

\begin{table}[!t]
\centering
\small
\setlength{\tabcolsep}{4pt}
\renewcommand{\arraystretch}{1.15}
\caption{Comparison of representative reward functions in RL-based agentic search. 
$a_{\text{pred}}$ and $a_{\text{gt}}$ denote the predicted and ground-truth answers, respectively. 
$r_{\text{ans}}$ is the answer-level reward; 
$RT$ is the number of retrieval steps; 
$RT_{\max}$ is the maximum retrieval budget; 
$r_{\text{kb}+}$ and $r_{\text{kb}-}$ denote the maximal knowledge-boundary reward and a small penalty, respectively. 
$\mathbb{I}(\cdot)$ is the indicator function, $\gamma$ the discount factor, $v(\cdot)$ the rollout value, and $\alpha$ a decay coefficient. 
$r_\text{sim}(\cdot,\cdot)$ is the reward function based on the semantic similarity between the
model-generated search query and the ground-truth query using a Sentence Transformer.
}
\label{tab:reward_comparison}
\vskip -1em
\begin{tabularx}{\textwidth}{
>{\RaggedRight\arraybackslash}X
>{\RaggedRight\arraybackslash}X
>{\RaggedRight\arraybackslash}X
>{\RaggedRight\arraybackslash}X
>{\RaggedRight\arraybackslash}p{5.3cm}
}
\toprule
\textbf{Reward Type} & \textbf{Method} & \textbf{RL Role} & \textbf{Reward Name} & \textbf{Reward Definition} \\
\midrule

\multirow{12}{*}{\textbf{Outcome}}
& Search-R1~\cite{jin2025searchr1} & Adapt-Search 
& Answer EM  
& $r = EM(a_{\text{pred}}, a_{\text{gt}})$ \\

\cmidrule(lr){2-5}
& ReSearcher~\cite{jin2025searchr1} & Adapt-Search 
& Answer F1  
& $r = F1(a_{\text{pred}}, a_{\text{gt}})$ \\

\cmidrule(lr){2-5}
& ReZero~\cite{dao2025rezero} & Search Intensity  
& Retry Reward 
& $\displaystyle
r =
\begin{cases}
\sum_{k=1}^{N_{\text{retry}}} \gamma^{k-1}, & \text{if format valid},\\[4pt]
0, & \text{otherwise.}
\end{cases}$ \\

\cmidrule(lr){2-5}
& Pangu DeepDiver~\cite{wang2025stepsearch} & Search Intensity 
& Extra Search Call Reward  
& $\displaystyle
r =
\begin{cases}
1, & \text{if uses search and answer is correct},\\[4pt]
0, & \text{otherwise.}
\end{cases}$ \\

\cmidrule(lr){2-5}
& IKEA~\cite{huang2025reinforced} & Search Efficiency
& Knowledge-Boundary-Aware Reward  
& \makecell[l]{%
$\displaystyle
r =
\begin{cases}
r_{\text{kb}+}\!\left(1 - \tfrac{RT}{RT_{\max}}\right), & r_{\text{ans}} = 1,\\[4pt]
0, & r_{\text{ans}} = 0 \land RT = 0,\\[4pt]
r_{\text{kb}-}, & r_{\text{ans}} = 0 \land RT > 0.
\end{cases}$%
} \\

\midrule
\multirow{10}{*}{\textbf{Process}}
& Autorefine~\cite{jin2025searchr1} & Retrieval–Search Interaction 
& Retrieval-Specific Reward  
& \makecell[l]{%
$\displaystyle 
r = \mathbb{I}\big(a_{\text{gt}} \cap a_{\text{refine}} = a_{\text{gt}}\big),$ where\\[4pt]
$\displaystyle 
a_{\text{refine}} = 
\bigcup_{t}\{\,c_t \mid (s_t, c_t)\!\in\! a,\, s_t = \langle\text{refine}\rangle\}.$%
} \\

\cmidrule(lr){2-5}
& R-Search~\cite{zhao2025r} & Retrieval–Search Interaction
& Evidence Quality Reward 
& \makecell[l]{%
$\displaystyle 
r = F1(\alpha_{\text{cf}}, \alpha_{\text{gold}}), \quad 
\alpha_{\text{cf}} \!\sim\! \pi_{\text{cf}}(\cdot \mid q,e)$%
} \\

\cmidrule(lr){2-5}
& ReasonRAG~\cite{zhang2025process} & Search Efficiency 
& Shortest-Path Reward Estimation  
& $\displaystyle
r = \tfrac{1}{h}\sum_{i=1}^{h} v(\text{rollout}_i)\cdot\alpha^{\text{step}(\text{rollout}_i)}$ \\

\cmidrule(lr){2-5}
& Visual-ARFT~\cite{liu2025visual} & \makecell[l]{Multi-Tool /\\Multi-Modal\\Adapt-Search}
& Semantic Similarity Reward  
& $\displaystyle 
r = r_\text{sim}(a_{\text{search}}, s)$ \\

\bottomrule
\end{tabularx}
\end{table}

\subsection{Reward Design}
\label{sec:reward_design}
Reward design is paramount in RL training for agentic search, determining which behaviors are reinforced and how credit is allocated across complex trajectories. Modern agentic search employs \emph{multi-faceted, multi-turn reward mechanisms} that optimize not only accuracy of final outcomes and intermediate reasoning, but also diverse desiderata such as clarity, truthfulness, conciseness, efficiency, and reduced hallucination tendencies. These sophisticated reward structures can be categorized along two complementary dimensions: temporal scope (outcome vs. process-level) and objective diversity (single vs. multi-faceted optimization). Table~\ref{tab:reward_comparison} summarizes representative reward functions adopted in recent RL-based agentic search frameworks~\cite{jin2025searchr1,chen2025learning,dao2025rezero,shi2025pangu,shi2025search,zhao2025r,zhang2025process,liu2025visual}, illustrating how different designs balance final-answer accuracy, intermediate reasoning quality, and resource-efficient retrieval.


\subsubsection{Outcome-level Rewards}
\label{sec:outcome_reward}
Outcome-level rewards evaluate final task completion but increasingly incorporate multiple quality dimensions beyond simple correctness.
Early approaches like Search-R1~\cite{jin2025searchr1} and ReSearch~\cite{chen2025learning} rely on basic exact match (EM) and format reward for correctness and style consistency. Subsequent \textbf{multi-faceted} extensions enhance these metrics: 
R-Search~\cite{zhao2025r} introduces \emph{cross-model evidence utility}, rewarding evidence quality and interpretability alongside correctness.
IKEA~\cite{huang2025reinforced} designs \emph{knowledge-boundary shaping} to optimize both accuracy and efficiency by discouraging redundant retrieval.
R1-Searcher++~\cite{song2025r1++} measure \emph{group-relative efficiency} through retriever call variance, balancing task success with resource conservation. 
O$^2$-Searcher~\cite{mei2025o2} introduces a \textit{diversity reward} to encourage \emph{query diversity} to mitigate duplication under budget constraints.

\subsubsection{Process-level Rewards}
\label{sec:process_reward}
While outcome signals are simple and effective for general tasks, they often prove too sparse to guide learning in long-horizon, multi-step search settings~\cite{deng2025atom}.
Process-level rewards address this limitation by providing dense, fine-grained feedback throughout the reasoning–retrieval trajectory, enabling \emph{multi-turn, multi-faceted} optimization of intermediate behaviors, such as faithfulness~\cite{shi2025search} and efficiency~\cite{wang2025stepsearch}.
ReasonRAG~\cite{zhang2025process} introduces \emph{shortest-path reward estimation} (SPRE), which simultaneously optimizes reasoning quality and conciseness by simulating its possible outcomes and penalizing unnecessarily long trajectories.
StepSearch~\cite{wang2025stepsearch} evaluates the utility of each retrieval step across multiple dimensions, including information gain and redundancy penalties.
AutoRefine~\cite{shi2025search} reinforces faithful and targeted knowledge extraction through iterative step-level rewards. 
In addition to these verifiable rule-based rewards, some works~\cite{wang2025sirag, deng2025atom} also sample rewards from LLMs for providing step-level rewards to address the sparse reward and training stability or enable faithful search~\citep{xu2025VERITAS}.


\begin{myboxi}
\textbf{Takeaways}:
\begin{itemize}[leftmargin=*]
    \item \textbf{Rewards have shifted from single-objective outcomes to multi-faceted objectives.}  
    Outcome-level signals now combine correctness with efficiency, interpretability, and diversity; process-level signals provide dense guidance (e.g., info-gain, redundancy penalties, shortest-path/length control, faithfulness).
    \item \textbf{Unifying outcomes and processes is key.}  
    Effective agents balance final accuracy with intermediate behavior quality; shaping should align step-wise improvements with end goals to avoid myopic optimization.
    \item \textbf{Open challenges.}  
    Credit assignment over long horizons, reward hacking/overfitting, objective balancing (accuracy–efficiency–faithfulness), and stable scaling (cost/latency) remain active problems; self-evolving loops (RL$\leftrightarrow$SFT) are promising but need careful control and evaluation.
\end{itemize}
\end{myboxi}

%% file: 6_Where_to_use_v2.tex
\section{Where RL is Applied: The Scope of Optimization}
\label{sec:where_RL_is_applied}

The application of RL in agentic search can be categorized by the \emph{architectural level} at which optimization occurs.  
This perspective clarifies whether RL refines specific sub-skills, optimizes the policy of a single agent, or orchestrates behavior across multi-agent or system-wide search infrastructures.  
We summarize representative works across these three levels of scope in Table~\ref{tab:where_to_use}.

\begin{table}[t]
\centering
\small

\setlength{\tabcolsep}{4pt}
\renewcommand{\arraystretch}{1.15}
\caption{The categorization of RL-based search agents from the optimization scope's perspective.}
\label{tab:where_to_use}
\vskip -1em
\begin{tabularx}{\textwidth}{
>{\RaggedRight\arraybackslash}p{1.8cm}
>{\RaggedRight\arraybackslash}p{3.5cm}
>{\RaggedRight\arraybackslash}X
}

\toprule
\textbf{Category} & \textbf{Optimization Scope} & \textbf{Methods} \\
\midrule

\multirow{11}{*}{\textbf{Agentic-level}}
&\multirow{10}{*}{Single-agent Optimization}
&
Search-R1\cite{jin2025searchr1}; ReSearch\cite{chen2025learning}; R1-Searcher++\cite{song2025r1++}; AutoCoA~\cite{zhang2025agent}; DeepRAG\cite{guan2025deeprag}; WebSailor\cite{li2025websailor}; WebSailor-V2~\cite{li2025websailorv2}; WebDancer\cite{wu2025webdancer}; WebThinker~\cite{li2025webthinker}; WebWatcher~\cite{geng2025webwatcher}; ExSearch~\cite{shi2025iterative}; GRAIL~\cite{chang2025grail}; DynaSearcher~\cite{hao2025dynasearcher}; SimpleDeepSearcher~\cite{sun2025simpledeepsearcher}; DeepResearcher\cite{zheng2025deepresearcher}; ReSum\cite{wu2025resum}; R-Search~\cite{zhao2025r}; ParallelSearch~\cite{zhao2025parallelsearch}; EvolveSearch~\cite{zhang2025evolvesearch}; O$^2$-Searcher~\cite{mei2025o2}; Pangu DeepDiver\cite{shi2025pangu}; IKEA~\cite{huang2025reinforced}; UR$^2$~\cite{li2025ur}; SSRL~\cite{fan2025ssrl}; ZeroSearch~\cite{sun2025zerosearch}; MaskSearch~\cite{wu2025masksearch}; ReZero~\cite{dao2025rezero}; Tool-Star~\cite{dong2025tool}; WebExplorer~\cite{liu2025webexplorer}; SFR-DeepResearch~\cite{nguyen2025sfr}; WebResearcher~\cite{qiao2025webresearcher};  Visual-ARFT~\cite{liu2025visual}; MMSearch-R1~\cite{wu2025mmsearch}; VRAG-RL~\cite{wang2025vrag}; Lucy~\cite{dao2025lucy}; MedResearcher-R1~\cite{yu2025medresearcher}; DeepRetrieval\cite{jiang2025deepretrieval};
Webthinker~\cite{li2025webthinker}
 \\
\cmidrule{2-3}
&
Multi-agent Coordination
&
HARIS~\cite{hu2025coordinating};  SIRAG\cite{wang2025sirag}; MAO-ARAG\cite{chen2025mao}; MMOA-RAG\cite{chen2025improving}; OPERA~\cite{liu2025opera}\\
\midrule

\multirow{3}{*}{\makecell[l]{\textbf{Module-Level}\\\textbf{\& Step-level}}}
&
Module-level Optimization
&
s3\cite{jiang2025s3}; AI-SearchPlanner\cite{mei2025ai}; DeepResearcher~\cite{zheng2025deepresearcher}
\\
\cmidrule{2-3}
&
\multirow{2}{*}{Step-level Optimization}
&
StepSearch\cite{wang2025stepsearch}; AutoRefine\cite{shi2025search}; Search Wisely\cite{wu2025search}; ConvSearch-R1~\cite{zhu2025convsearch}; Atom-Searcher~\cite{deng2025atom}; ReasonRAG\cite{zhang2025process}; SWiRL~\cite{goldie2025synthetic}; Atom-Searcher~\cite{deng2025atom}; \\
\midrule

\multirow{2}{*}{\textbf{System-level}}
&
Unified RL-based Agentic Framework
&
AgentGym-RL\cite{xi2025agentgym}; Verl~\cite{sheng2024hybridflow}; VerlTool\cite{jiang2025verltool}; RAG-Gym\cite{xiong2025rag}; Chain-of-Agents\cite{li2025chain} \\
\bottomrule
\end{tabularx}

\end{table}

\subsection{Agent-level Scpoe}
\label{sec:agent_level_scope}
At the agent level, RL optimizes end-to-end search policies, either for single autonomous search agents or coordinated multi-agent search systems. This scope captures how RL shapes the core search decision-making processes that define effective information-seeking behavior.

\subsubsection{Single-agent Optimization}
\label{sec:single_agent}
This is the most prevalent paradigm, where RL directly optimizes a unified policy governing the agent’s entire search workflow.  
The agent learns when to retrieve, how to formulate queries, how to interpret evidence, and when to terminate its search.  
Search-R1~\cite{jin2025searchr1} exemplifies this approach, training an LLM to autonomously decide when and how to invoke external search engines during reasoning.  
R1-Searcher++~\cite{song2025r1++} extends this by balancing internal knowledge use with external search reliance.  
Web-based agents such as WebSailor~\cite{li2025websailor} and WebDancer~\cite{wu2025webdancer} demonstrate RL’s potential to train robust, long-horizon search policies for complex web environments.

\subsubsection{Multi-agent Coordination}
\label{sec:multi_agent_corrdination}
For more complex search pipelines, distinct agents specialize in  search-related functions such as query reformulation, document selection, and evidence synthesis.  
RL coordinates these specialized search agents to achieve coherent information-seeking behavior. 
SIRAG~\cite{wang2025sirag} jointly trains a \emph{Decision Maker} to control search timing and a \emph{Knowledge Selector} to filter retrieved documents under a shared reward function.  
MAO-ARAG~\cite{chen2025mao} orchestrates multiple search specialists (e.g., query reformulators, document selectors, answer generators) using RL to optimize their collaborative search performance.


\subsection{Module-Level \& Step-level Scope}
\label{sec:module_level_step_level}
This scope focuses on optimizing specific search components or decision steps within broader agentic search workflows. Instead of training the entire agent policy end-to-end, RL refines localized behaviors, making it valuable for improving specific aspects of the search pipeline.

\subsubsection{Module-level Optimization} 
\label{sec:module_level_opt}
RL can enhance specialized modules that operate alongside frozen LLMs.  
This modular approach isolates search-specific capabilities for targeted improvement without full-model retraining.  
The s3~\cite{jiang2025s3} exemplifies this strategy by training a lightweight searcher module while keeping the generator frozen, ensuring efficiency and model-agnostic adaptability.  
AI-SearchPlanner~\cite{mei2025ai} follows a similar design, training a retrieval-planning module to decide when and how to query while leveraging a frozen QA model for final answer generation.

\subsubsection{Step-level Optimization}
\label{sec:step_level_opt}
RL can also provide fine-grained feedback on individual search actions, such as query generation, document selection, or refinement.  
StepSearch~\cite{wang2025stepsearch} provides step-wise rewards based on information gain and redundancy penalties to encourage concise, effective search.  
AutoRefine~\cite{shi2025search} reinforces iterative “search-and-refine” behaviors, encouraging agents to iteratively improve their information gathering.  
Search Wisely~\cite{wu2025search} applies RL to control retrieval confidence, discouraging low-confidence searches that waste resources.

\subsection{System-level Scope}
\label{sec:system_level}
At the system level, RL orchestrates comprehensive search infrastructures and multi-agent search ecosystems. Rather than optimizing individual search agents, this scope addresses how RL can improve entire search system architectures, resource allocation, and search workflow management across complex information-seeking platforms.

\subsubsection{Unified RL-based Framework for Search}
\label{sec:unified_RL-based_framework}
Several recent works build general-purpose platforms for developing, training, and evaluating RL-based search agents.  
AgentGym-RL~\cite{xi2025agentgym} provides a modular benchmark suite that supports diverse RL algorithms across multiple information environments.  
RAG-Gym~\cite{xiong2025rag} offers structured environments for optimizing retrieval-augmented agents and systematically comparing reward and policy designs.  
VerlTool~\cite{jiang2025verltool} extends this trend to tool-augmented systems, offering unified APIs and environments for training agents that operate over heterogeneous information sources and modalities.




\begin{myboxi}
\textbf{Takeaways}:
\begin{itemize}[leftmargin=*]
    \item \textbf{Agent-level RL} establishes the foundation for end-to-end search intelligence.  
    Single-agent optimization yields coherent policies for when and how to search, while multi-agent coordination introduces modular specialization and interpretability.  
    The trade-off lies between unified autonomy and orchestrated collaboration.

    \item \textbf{Module- and step-level RL} provide fine-grained control for improving local behaviors without full-model retraining.  
    Module-level tuning enhances efficiency via lightweight plug-ins, and step-level rewards supply dense supervision for precise search decisions.  
    However, effective \emph{credit assignment} remains an open challenge for connecting local improvements to global task success.

    \item \textbf{System-level RL} extends beyond individual agents to entire infrastructures.  
    Frameworks such as AgentGym-RL and RAG-Gym foster reproducibility, standardized evaluation, and scalable experimentation—marking a shift from isolated prototypes to deployable, ecosystem-level optimization.

    \item \textbf{Across levels}, the scope of RL optimization reflects a continuum:  
    from micro-level behavioral refinement, through agent-level policy learning, to macro-level system orchestration.  
    Future progress will hinge on unifying these layers—developing hierarchical or multi-scale RL frameworks that integrate step-wise feedback, agent collaboration, and system-wide efficiency under shared reward principles.
\end{itemize}
\end{myboxi}


%% file: 7_Evaluation.tex
\section{Evaluation and Application}
\label{sec:evaluation_application}
Evaluating RL-based agentic search systems requires multi-dimensional assessment across search effectiveness, reasoning quality, efficiency, and generalization.  
This section reviews the datasets, evaluation metrics, and application domains that currently define the landscape of RL-based agentic search evaluation and deployment.

\begin{table}[t]
\centering
\small
\setlength{\tabcolsep}{4pt}
\renewcommand{\arraystretch}{1.15}
\caption{The categorization of commonly used datasets in RL-based agentic search.}
\label{tab:dataset}
\vskip -1em
\begin{tabularx}{\textwidth}{
>{\RaggedRight\arraybackslash}p{3.0cm}
>{\RaggedRight\arraybackslash}X
}

\toprule \textbf{Category} & \textbf{Dataset} \\
\midrule

Knowledge Source & wiki-dump~\cite{wikimedia_dumps}; Common Crawl~\cite{commoncrawl_overview}; KILT~\cite{petroni2020kilt}; PubMed~\cite{pubmed_about}; Arxiv~\cite{arxiv_about};
 \\
\midrule
\multirow{5}{*}{Knowledge-Intensive QA} & 
NQ~\cite{kwiatkowski2019natural};
TriviaQA~\cite{joshi2017triviaqa}; HotpotQA~\cite{yang2018hotpotqa}; 2WikiMultiHopQA~\cite{ho2020constructing}; MuSiQue~\cite{trivedi2022musique}; PopQA~\cite{mallen2022not}; CAG~\cite{pan2024cag}; C-SimpleQA~\cite{wei2024measuring}; 
SuperGPQA~\cite{du2025supergpqa};
BRIGHT~\cite{su2024bright}; SealQA~\cite{pham2025sealqa}; BLUR~\cite{ch2025browsing};
NaturalReasoning~\cite{yuan2025naturalreasoning}
FEVER~\cite{thorne2018fever}; EX-FEVER~\cite{ma2024ex}; FEVEROUS~\cite{aly2021feverous}; FactBench~\cite{bayat2024factbench}; RealFactBench~\cite{yang2025realfactbench}; LongFact~\cite{wei2024long};
FRAMES~\cite{krishna2024frames}
RAG-Bench~\cite{friel2024ragbench};
BEIR~\cite{thakur2021beir}; AmbigQA~\cite{min2020ambigqa}; MetaQA~\cite{puerto2021metaqa};WebQuestions~\cite{berant2013webquestions};
CWQ~\cite{talmor2018web}; CheckWhy~\cite{si2024checkwhy}; BeerQA~\cite{qi2021answering}\\
\midrule
\multirow{3}{*}{Web-based Search} & WebQA~\cite{chang2022webqa}; Bamboogle~\cite{press2022measuring}; Mind2Web~\cite{gou2025mind2web}; 
WebArena~\cite{zhou2023webarena};
WebWalkerQA~\cite{wu2025webwalker};
AgentBench~\cite{liu2024agentbench};
BrowseComp-en~\cite{bc_en}; BrowseComp-zh~\cite{bc_zh}; GAIA~\cite{mialon2023gaia}; GAIA-2~\cite{russell2025gaia2};
XbenchDeepSearch~\cite{xbench}; WebPuzzle~\cite{shi2025pangu}; InfoDeepSeek~\cite{xi2025infodeepseek}; ORION~\cite{huang2025manusearch}; WebShaperQA~\cite{tao2025webshaper} \\
\midrule
\multirow{3}{*}{Multi-modal} & InfoSeek~\cite{chen2023can}; MMSearch~\cite{jiang2024mmsearch}; MMSearch-Plus~\cite{tao2025mmsearch} SimpleVQA~\cite{cheng2025simplevqa}; LiveVQA~\cite{fu2025livevqa}; MM-BrowseComp~\cite{li2025mm}; MAT-Search~\cite{liu2025matsearch}; Mocheg~\cite{yao2023end}; MFC-Bench~\cite{wang2024mfc}; ViDoSeek~\cite{wang2025vidorag}; SlideVQA~\cite{tanaka2023slidevqa}; MMLongBench~\cite{ma2024mmlongbench}\\ 
\midrule
\multirow{2}{*}{Conversational} & CoQA~\cite{reddy2019coqa}; QuAC~\cite{choi2018quac}; MSMarco~\cite{bajaj2016ms}; TopiOCQA~\cite{adlakha2022topiocqa}; QReCC~\cite{anantha2021open}; OR-QuAC~\cite{qu2020open}; NarrativeQA~\cite{kovcisky2018narrativeqa}; Doc2Dial~\cite{feng2020doc2dial}\\
\midrule
\multirow{5}{*}{Domain-specific }& 
MATH~\cite{hendrycks2021measuring};
MATH500~\cite{lightman2023let}
AIME24~\cite{aime2024};
AIME25~\cite{aime2025};
GSM8K~\cite{cobbe2021training};
Minerva~\cite{lewkowycz2022solving};
MMLU~\cite{hendrycks2020measuring}; 
MMLU-Pro~\cite{wang2024mmlu};
NuminaMath~\cite{li2024numinamath};
MedQA~\cite{jin2021disease};
MedMCQA~\cite{pal2022medmcqa}; MedBrowseComp~\cite{chen2025medbrowsecomp}
OlympiadBench~\cite{he2024olympiadbench}; USACO~\cite{shi2024can}; HLE~\cite{phan2025humanity}
FinSearchBench-24~\cite{li2024agent}; FinAgentBench~\cite{choi2025finagentbench} xbench~\cite{chen2025xbench}; 
MIRAGE~\cite{dongre2025mirage}; SolutionBench~\cite{li2025deepsolution}; DQA~\cite{lee2024planrag}; 
AirQA~\cite{huang2025airqa};
HERB~\cite{choubey2025deepsearch};
SciQ~\cite{welbl2017crowdsourcing}; SciFact~\cite{wadden2020fact};ARC~\cite{clark2018think}; ScIRGen-Geo~\cite{lin2025scirgen};
DeepShop~\cite{lyu2025deepshop};
NFCorpus~\cite{boteva2016};
OpenThoughts~\cite{openthoughts2025open};
\\

\bottomrule
\end{tabularx}

\end{table}

\subsection{Datasets}
\label{sec:datasets}
RL-based agentic search is evaluated across diverse benchmarks that test retrieval effectiveness and reasoning ability in open-domain, web-based, and domain-specific settings. Table~\ref{tab:dataset} summarizes these representative datasets and the corresponding studies that adopt them. Next, we give the details.

\subsubsection{Knowledge-Intensive QA Benchmarks}
\label{sec:knowledge_intensive_qa}
A primary evaluation setting for agentic search is \emph{knowledge-intensive question answering (QA)}, where answering a question requires retrieving external evidence beyond the model’s parametric knowledge. These benchmarks jointly evaluate the agent’s ability to (i) retrieve relevant information and (ii) synthesize evidence into correct, verifiable answers.  
Natural Questions (NQ)~\cite{kwiatkowski2019natural} and TriviaQA~\cite{joshi2017triviaqa} serve as foundational single-hop QA datasets, widely used in works such as Search-R1~\cite{jin2025searchr1} and R-Search~\cite{zhao2025r}, to test when and how agents invoke retrieval.  
For multi-hop reasoning, HotpotQA~\cite{yang2018hotpotqa} is employed in ReSearch~\cite{chen2025learning} and AutoRefine~\cite{shi2025search}, requiring iterative retrieval and reasoning over multiple evidence chains.  
Fact-checking tasks such as FEVER~\cite{thorne2018fever} further test retrieval faithfulness and evidence verification. HARIS~\cite{hu2025coordinating}, for instance, uses FEVER to train agents that assess the credibility of retrieved claims under RL signals.  

\subsubsection{Web-based Search Benchmarks}
\label{sec:web_based_search_bench}
Web environments provide more realistic and dynamic evaluation settings.  
WebQA~\cite{chang2022webqa} offers large-scale web-based QA tasks used in WebThinker~\cite{li2025webthinker}.  
GAIA (General AI Assistant) defines multi-step, interactive web tasks requiring reasoning and tool coordination, serving as a key benchmark for AgentGym-RL~\cite{xi2025agentgym} and WebSailor-V2~\cite{li2025websailorv2}.  
Mind2Web~\cite{gou2025mind2web} and related web navigation datasets evaluate the ability of web agents such as WebDancer~\cite{wu2025webdancer} to handle multi-hop web browsing and action planning.


\subsubsection{Knowledge Sources}
\label{sec:knowledge_sources}
Most open-domain and web-based agents rely on large-scale text corpora as retrieval backends. Common choices include the English Wikipedia dump~\cite{wikimedia_dumps}, widely used in benchmarks such as NQ, TriviaQA, and HotpotQA; web-scale resources such as Common Crawl~\cite{commoncrawl_overview} and KILT~\cite{petroni2020kilt}; and domain-specific knowledge bases such as PubMed~\cite{pubmed_about} and arXiv~\cite{arxiv_about}, which support research-oriented agents~\cite{zheng2025deepresearcher,yu2025medreseacher}. Some systems, including DeepResearcher~\cite{zheng2025deepresearcher} and WebThinker~\cite{li2025webthinker}, further augment these static corpora with dynamic web-search APIs to access up-to-date or domain-targeted information.

\subsubsection{Multi-modal Search}
\label{sec:dataset_multi_modal_search}
Recent advances in agentic search~\cite{wu2025mmsearch,liu2025visual} extend beyond text-only retrieval to incorporate visual and structured modalities, motivating new benchmarks for \emph{multi-modal search}.  
Early datasets, e.g., \textbf{InfoSeek}~\cite{chen2023can} and \textbf{SlideVQA}~\cite{tanaka2023slidevqa}, established vision–language question answering over slides and figures, bridging perception and reasoning.  
Building on this foundation, \citet{liu2025visual} introduce \emph{MAT-Search} and \emph{MAT-Coding} to evaluate agentic retrieval and tool-use abilities under verifiable reward signals.  
\textbf{MFC-Bench}~\cite{wang2024mfc} benchmarks multimodal fact-checking with $35k$ image–text samples across manipulation, out-of-context, and veracity subtasks, providing a large-scale testbed for factual grounding.  
Meanwhile, \textbf{MMLongBench-Doc}~\cite{ma2024mmlongbench} focuses on long-context multimodal document understanding, covering $135$ lengthy documents that combine text, layout, tables, and charts.  
Together, these benchmarks advance RL-based agentic search toward unified, perception-grounded multi-modal retrieval and reasoning.

\subsubsection{Conversational and Multi-turn Search}
\label{sec:conversational}
CoQA~\cite{reddy2019coqa} and QuAC~\cite{choi2018quac} benchmark the ability of agents to maintain context across multi-turn interactions, as explored in ConvSearch-R1~\cite{zhu2025convsearch}.  
MSMarco~\cite{bajaj2016ms} evaluates large-scale passage retrieval and ranking, assessing an agent’s ability to locate relevant information efficiently, as applied in DeepRetrieval~\cite{jiang2025deepretrieval} and RAG-Gym~\cite{xiong2025rag}.


\subsubsection{Domain-specific Search Tasks}
\label{sec:domain_specific}
Some specialized datasets~\cite{welbl2017crowdsourcing, clark2018think, talmor2018commonsenseqa, hendrycks2020measuring} target specific reasoning domains. For instance, SciQ~\cite{welbl2017crowdsourcing} and ARC~\cite{clark2018think} focus on scientific reasoning, relevant to agents like DeepResearcher~\cite{zheng2025deepresearcher}.  
CommonsenseQA~\cite{talmor2018commonsenseqa} tests the integration of factual retrieval and commonsense reasoning, used in IKEA~\cite{huang2025reinforced}.  
MMLU~\cite{hendrycks2020measuring} evaluates general knowledge breadth, serving as a multi-domain benchmark for tool-augmented systems such as Tool-Star~\cite{dong2025tool}.


\subsection{Metrics}
\label{sec:metric}
Evaluating RL-based agentic search requires metrics that capture multiple dimensions of performance, including answer quality, retrieval effectiveness, efficiency, and process-level behavior.  

\subsubsection{Answer Quality}
\label{sec:answer_quality}
Exact Match (EM) and F1 score are two of the most commonly used metrics, which provide direct measures of task success, serving as primary evaluation metrics in many works~\cite{jin2025searchr1,dao2025rezero}. To evaluate the generated answer quality against reference responses, ROUGE and BLEU scores evaluate generated answer quality against reference responses. To handle the case that answers may be correct but phrased differently from gold standards, BERTScore~\cite{zhang2019bertscore} is applied in RAG-Gym~\cite{xiong2025rag}.

\subsubsection{Search Effectiveness}
\label{sec:search_effectiveness}
To measure the quality of the retrieved information, several traditional information retrieval metrics remain fundamental. Specifically, \emph{Precision}, \emph{Recall}, and \emph{F1} measure the quality of retrieved information. Mean Reciprocal Rank (MRR) and Normalized Discounted Cumulative Gain (NDCG) evaluate ranking quality when systems need to prioritize multiple search results. For example, DeepRetrieval~\cite{jiang2025deepretrieval} trains LLMs to generate queries that maximize the retrieval performance of black-box search engines in terms of retrieval metrics like Recall and NDCG.

\subsubsection{Search Efficiency}
\label{sec:metric_search_efficiency}
It aims to measure search agents' efficiency from both resource and latency cost perspectives. \emph{Number of Search Queries}~\cite{shi2025pangu} measures how many queries an agent issues, while \emph{API Call Cost}~\cite{chen2025mao} quantifies the expense of invoking external services. \emph{Response Time} assesses end-to-end latency, important for interactive settings. \emph{Search Redundancy}~\cite{song2025r1++} captures repeated or semantically similar queries that waste resources.

\subsubsection{Process Metrics}
\label{sec:process_metric}
Beyond end-task accuracy, several works assess intermediate behaviors.  
StepSearch~\cite{wang2025stepsearch} defines \emph{Information Gain} per retrieval step to quantify the utility of each search action.  
SIRAG~\cite{wang2025sirag} measures \emph{Query Quality Score} via LLM-as-Judge to evaluate whether generated queries are likely to yield relevant evidence.  
R-Search~\cite{zhao2025r} introduces \emph{Evidence Utilization Rate} to measure how effectively agents leverage retrieved information in final reasoning.



\subsection{Applications}
\label{sec:applications}
The progress in RL-based agentic search has led to broad practical applications spanning scientific research, software development, multi-modal reasoning, and conversational AI.  


\subsubsection{Deep Research}
\label{sec:application_deep_resarch}
Scientific and academic research represents a major application domain for RL-based search agents. DeepResearcher~\cite{zheng2025deepresearcher} demonstrates automated literature review and hypothesis generation through RL-optimized search strategies across academic databases. MedResearcher-R1~\cite{yu2025medreseacher} specializes in medical research, using RL to navigate complex biomedical knowledge bases and synthesize clinical evidence. WebResearcher~\cite{qiao2025webresearcher} extends research capabilities to general web-based investigation with unbounded reasoning horizons. SFR-DeepResearch~\cite{nguyen2025sfr} focuses on autonomous reasoning for research tasks, while Atom-Searcher~\cite{deng2025atom} enhances deep research through fine-grained atomic thought rewards. WebThinker~\cite{li2025webthinker} is a deep research agent empowered with comprehensive research capabilities across diverse domains through iterative online DPO.

\subsubsection{Multi-modal Search}
\label{sec:application_multi_modal}
In addition to text-only search, there are several recent efforts~\cite{wu2025mmsearch,wang2025vrag} exploring multi-modality search agents, combining both text and visual information. VRAG-RL~\cite{wang2025vrag} enables vision-perception-based RAG for visually rich information understanding, using RL to iteratively reason across both textual and visual content. Visual-ARFT~\cite{liu2025visual} demonstrates visual agentic reinforcement fine-tuning for tasks requiring integrated visual and textual search. WebWatcher~\cite{geng2025webwatcher} breaks new ground in vision-language deep research agents, combining web search with visual analysis capabilities. These applications are particularly valuable in domains like e-commerce, where product search requires understanding both descriptions and images, and in scientific research involving visual data analysis.

\subsubsection{Code Agents}
\label{sec:application_code_agent}
Beyond typical search-related applications, RL-powered search agents are being integrated into programming and software development workflows. Tool-Star~\cite{dong2025tool} demonstrates multi-tool reasoning capabilities that include code execution and debugging, using RL to coordinate between search engines, code interpreters, and other development tools.
VerlTool~\cite{jiang2025verltool} provides a unified framework for agentic RL with tool use that specifically supports code interpreters alongside other APIs, enabling agents to search for code solutions, execute them, and iteratively refine implementations. These systems learn to balance web search for coding solutions with direct code experimentation, optimizing both information gathering and implementation efficiency.

\subsubsection{AI Assistants}
\label{sec:application_ai_assistant}
Conversational AI is a growing deployment area for RL-based search agents, which is far beyond a naive chatbot but like a personal assistant with the capability to handle various realistic tasks. For instance, ConvSearch-R1~\cite{zhu2025convsearch} specifically addresses conversational search scenarios, using RL to enhance query reformulation and maintain context across multi-turn interactions. Lucy~\cite{dao2025lucy} demonstrates edge-running agentic web search on mobile devices with machine-generated task vectors, showcasing practical deployment in resource-constrained environments. MAO-ARAG~\cite{chen2025mao} provides adaptive retrieval-augmented generation through multi-agent orchestration, suitable for intelligent assistant applications that need to balance response quality with computational efficiency. 
These systems use RL to learn to understand user intent, search for relevant information, and provide contextually appropriate responses while maintaining conversation flow.

\subsubsection{Domain-specific Applications}
\label{sec:application_domain_specific}
In addition to the aforementioned general applications, RL-based search agents are also applied in specialized domains tailored to specific knowledge areas and user needs. For instance,
HierSearch~\cite{tan2025hiersearch} presents enterprise search frameworks that integrate local knowledge bases with web search, addressing corporate information management needs. 
KunLunBaizeRAG~\cite{li2025kunlunbaizerag} focuses on inference performance optimization for large language models in domain-specific RAG scenarios.
DynaSearcher~\cite{hao2025dynasearcher} demonstrates dynamic knowledge graph (KG) augmented search for structured information retrieval, particularly valuable in domains with rich relational data.
GRAIL~\cite{chang2025grail} enables interactive KG exploration for retrieval-augmented reasoning through RL.

\subsubsection{Takeaways}
The diversity of applications demonstrates the broad applicability and practical value of RL-based agentic search systems. From code development~\cite{dong2025tool} to scientific research~\cite{zheng2025deepresearcher}, multi-modal understanding~\cite{wu2025mmsearch}, conversational AI~\cite{zhu2025convsearch}, and specialized domains~\cite{tan2025hiersearch}, these systems address real-world information-seeking challenges across multiple sectors. The success of these applications highlights the importance of domain-specific adaptation, multi-modal capabilities, and efficient resource management in practical deployments. 
Future applications will likely see increased integration across modalities and domains, with RL enabling agents to adapt their search strategies dynamically based on task requirements and user contexts.

%% file: 8_Future_direction_v2.tex
\section{Challenges and Future Directions}
\label{sec:challenge_future_direction}
Despite the remarkable strides of RL-based agentic search, many fundamental challenges and opportunities lie ahead. In this section, we discuss key future directions that will shape the evolution of intelligent search agents, addressing both technical limitations and emerging requirements for real-world deployment.

\noindent\textbf{Multi-modal Agentic Search}. 
Real-world information exists across multiple modalities, including text, images, videos, audio, and structured data. Current RL-based search agents primarily focus on textual information, limiting their applicability to complex, multi-modal information-seeking tasks that require understanding and reasoning across diverse content types. While initial efforts~\cite{wu2025mmsearch,wang2025vrag,geng2025webwatcher} enable search engines to facilitate reasoning in vision-language models~\cite{bordes2024introduction,gao2025pixels,wu2025image}, several fundamental limitations persist: (i) how to ensure consistency between textual descriptions and visual content during search-integrated reasoning; (ii) how to determine which modality contributes most to successful outcomes in multi-modal search tasks; and (iii) how to design reward functions that jointly capture relevance, coherence, and cross-modal alignment. Addressing these challenges is essential for moving toward robust multi-modal agentic search, where agents can adaptively select, integrate, and reason over heterogeneous sources to solve open-ended real-world queries.

\noindent\textbf{Memory-augmented and Long-horizon Search}. 
Real-world information-seeking often spans multiple sessions, where agents must remember past queries, retrieved evidence, or user feedback. Current RL-based search agents~\cite{jin2025searchr1,zhao2025r} typically operate within limited context windows and lack sophisticated memory mechanisms for long-term information retention and retrieval. While some initial efforts~\cite{nguyen2025sfr,wu2025resum} consider simple memory management techniques such as summarization and cleanup operations, they still struggle with more complex tasks requiring long-term interactions and cross-session continuity. To advance agentic search in long-horizon scenarios, future research should explore developing sophisticated memory architectures that can selectively store, organize, and retrieve search-related knowledge over time. 
Promising directions include: (i) \emph{hierarchical memory systems} that differentiate between short-term working memory, episodic memory across sessions, and long-term semantic knowledge; (ii) \emph{selective memory} mechanisms that use RL signals to decide what retrieved information to retain, compress, or discard based on long-term utility; and (iii) \emph{temporal reasoning integration} that allows agents to model information decay, relevance shifts, and evolving user intents; 


\noindent\textbf{Trustworthy Agentic Search}.
Search agents operating in open environments face pressing security, ethical, and reliability challenges that directly affect user trust. These agents may encounter adversarial content, misinformation, or malicious actors attempting to manipulate their behavior for harmful purposes. Existing studies have revealed significant vulnerabilities in search-augmented systems. For instance, PoisonedRAG~\cite{zou2025poisonedrag} demonstrates that RAG can be misled by injected malicious knowledge, resulting in incorrect or unsafe outputs. While Search Wisely~\cite{wu2025search} explores uncertainty-aware search to mitigate overconfidence, it remains unclear how search agents perform under adversarial conditions and how to guarantee robustness in real-world deployments. Moreover, these agents frequently interact with sensitive information, raising concerns about privacy protection, ethical information use, and compliance with data governance regulations. Future research should investigate how to develop reliable, privacy-preserving and ethically aligned search agents. Promising directions include: (i) \emph{adversarially robust RL training}, where agents are exposed to poisoned or noisy retrieval environments to learn resilient policies; (ii) \emph{privacy-preserving agentic search}, such as federated or encrypted search agents, to safeguard sensitive user information; (iv) \emph{value-aligned reward design}, ensuring that optimization objectives incorporate fairness, transparency, and safety constraints; and (v) \emph{auditing and verification tools} that allow both developers and end users to interpret, monitor, and evaluate agent behavior. In conclusion, these approaches would move RL-based agentic search toward systems that are not only effective but also secure, ethical, and trustworthy for real-world applications.

\noindent\textbf{Cross-domain Generalization}. 
Current RL-based search agents are often trained for specific domains or tasks, limiting their generalizability. Real-world deployment requires agents that can adapt their search strategies across diverse domains and contexts.
To solve this challenge and expand agentic search to broader applications, future works can focus on learning generalizable search principles that can be applied across diverse contexts. For example, one potential solution is to develop meta learning approach to to create universal search strategies that can transfer across different information spaces, or to build agents that can automatically identify and adapt to domain-specific search requirements. 

\noindent\textbf{Human–AI Co-search}.
Traditional IR systems were designed for humans as the primary end users~\cite{marchionini2006exploratory,white2009exploratory}. The integration of retrieval into large-scale AI systems has reshaped this paradigm, particularly with the rise of LLMs. Retrieval is no longer performed solely for human consumption but increasingly serves to enhance models’ reasoning and generation capabilities~\cite{xu2025survey}. This shift raises fundamental questions about \textit{how humans and AI agents will collaboratively engage in exploratory search}. RL–based agentic search systems provide a natural foundation for this shift. Through interaction and feedback, RL enables agents to learn adaptive retrieval policies that align with evolving user intents and contextual cues, fostering \emph{human–AI co-search} where agents act as copilots that assist users in locating, interpreting, and synthesizing information.
Future research may explore:
(i) \textit{Adaptive interaction modeling}, where RL agents learn user preferences and search behaviors to personalize strategies and result presentation;
(ii) \textit{Explainable search reasoning}, allowing agents to justify retrieval choices and promote transparency;
(iii) \textit{Collaborative query refinement}, enabling iterative reformulation of search goals through natural-language interaction.


%% file: 9_Conclusion.tex
\section{Conclusion} 
The integration of RL into agentic search marks a fundamental shift in how LLMs interact with external knowledge. Unlike naive RAG, RL enables agents to dynamically decide \emph{when}, \emph{what}, and \emph{how} to search, transforming search into an adaptive and interactive process. This survey provides the first systematic overview of RL-based agentic search, synthesizing research across three perspectives: (i) \emph{What RL is for}; (ii) \emph{How RL is used}; and (iii) \emph{Where RL is applied}. We further examine evaluation metrics, system benchmarks, and representative applications, offering a comparative view of current progress. Looking ahead, RL-based agentic search holds the potential to redefine information retrieval and reasoning. We hope this survey provides a foundation for advancing research in this emerging field and inspires new directions toward practical, robust, and intelligent agentic search systems.

\begin{center}
{
\scriptsize
\begin{longtable}{
>{\RaggedRight\arraybackslash}m{1.6cm}
>{\RaggedRight\arraybackslash}m{1.8cm}
>{\RaggedRight\arraybackslash}m{0.6cm}
>{\RaggedRight\arraybackslash}m{1.0cm}
>{\RaggedRight\arraybackslash}m{1.2cm}
>{\RaggedRight\arraybackslash}m{1.2cm}
>{\RaggedRight\arraybackslash}m{1.8cm}
>{\RaggedRight\arraybackslash}m{1.8cm}
>{\RaggedRight\arraybackslash}m{1.6cm}
}
\caption{Overview of RL-based agentic search from the perspective of reinforcement learning optimization strategies. {ORM} and {PRM} denote the \emph{Outcome Reward Model} and the \emph{Process Reward Model}, respectively. “Rule-based” indicates that the reward function is entirely computed from predefined rules; otherwise, an LLM is involved as a reward judge.}
\label{tab:how_to_use} \\ 
\toprule
\textbf{Method} & \textbf{RL Func. Role} & \textbf{Cold Start?} & \textbf{Training Env.} & \textbf{RL Alg.} & \textbf{Reward Type} & \textbf{Reward Func.} & \textbf{Opt. Scope} & \textbf{Dataset}\\
\midrule
\endfirsthead
\toprule
\textbf{Method} & \textbf{RL Func. Role} & \textbf{Cold Start?} & \textbf{Training Env.} & \textbf{RL Alg.} & \textbf{Reward Type} & \textbf{Reward Func.} & \textbf{Opt. Scope} & \textbf{Dataset}\\
\midrule
\endhead
Search-R1~\cite{jin2025searchr1} & Adapt-Search  & \xmark & Real-world & \makecell[l]{PPO \\ GRPO} & Rule-based ORM & Answer EM & Single-agent & \cite{kwiatkowski2019natural,joshi2017triviaqa,mallen2022not,yang2018hotpotqa,trivedi2022musique,ho2020constructing,press2022measuring} \\
\midrule
ReSearch~\cite{chen2025learning} & Adapt-Search  & \xmark & Real-world & GRPO & Rule-based ORM &\makecell[l]{Format\\Answer F1} & Single-agent & \cite{yang2018hotpotqa,ho2020constructing,trivedi2022musique,press2022measuring} \\
\midrule
AutoCoA~\cite{zhang2025agent} & Adapt-Search & \cmark & Real-world & GRPO & Rule-based ORM & \makecell[l]{Format\\Answer EM} & Single-agent & \cite{kwiatkowski2019natural,joshi2017triviaqa,mallen2022not,yang2018hotpotqa,trivedi2022musique,ho2020constructing,press2022measuring}\\
\midrule
\makecell[l]{SimpleDeep-\\Searcher~\cite{sun2025simpledeepsearcher}} & Adapt-Search & \cmark & Real-world & \makecell[l]{DPO\\ Reinforce++} & Rule-based ORM & \makecell[l]{Format\\Answer F1} & Single-agent & \cite{yang2018hotpotqa,ho2020constructing,trivedi2022musique,tang2024multihop,krishna2024frames,press2022measuring,mialon2023gaia,zhou2025browsecomp,wei2025browsecomp}\\
\midrule
ExSearch~\cite{shi2025iterative} & Adapt-Search & \xmark & Real-world & GEM & PRM & \makecell[l]{Trajectory Quality} & Single-agent & \cite{kwiatkowski2019natural,yang2018hotpotqa,trivedi2022musique} \\
\midrule
IKEA~\cite{huang2025reinforced} & Search Efficiency & \xmark & Real-world & GRPO & Rule-based ORM & \makecell[l]{Format\\Answer EM\\Knowledge-boundary} & Step-level & \cite{kwiatkowski2019natural,mallen2022not,yang2018hotpotqa,ho2020constructing}\\
\midrule
R1-Searcher~\cite{song2025r1} & Adapt-Search & \cmark & Real-world & \makecell[l]{GRPO\\Reinforce++} & Rule-based ORM & \makecell[l]{Format\\Answer F1} & Single-agent &\cite{yang2018hotpotqa,ho2020constructing,press2022measuring,trivedi2022musique} \\
\midrule
R1-Searcher++~\cite{song2025r1++} & \makecell[l]{Search Efficiency} & \cmark & Real-world & \makecell[l]{GRPO\\Reinforce++} & Rule-based ORM & \makecell[l]{Format\\Answer EM\\Std of Search Calls}& Single-agent & \cite{yang2018hotpotqa,ho2020constructing,press2022measuring,trivedi2022musique}\\
\midrule
DeepRAG~\cite{guan2025deeprag} & \makecell[l]{Adapt-Search\\Search Efficiency}  & \cmark & Real-world & GRPO & Rule-based ORM & \makecell[l]{Answer EM \\ Retrieval Cost}& Single-agent & \cite{yang2018hotpotqa,ho2020constructing,pan2024cag,mallen2022not,berant2013webquestions,trivedi2022musique}\\
\midrule
UR$^2$~\cite{li2025ur} & Adapt-Search & \cmark & \makecell[l]{Real-world\\Curriculum} & Reinforce++ & Rule-based ORM & \makecell[l]{Format \\ Answer EM \\ Fallback Penalty} & Single-agent & \cite{hendrycks2021measuring,lewkowycz2022solving,jin2021disease,wang2024mmlu,yang2018hotpotqa,press2022measuring,ho2020constructing,trivedi2022musique}\\
\midrule
SSRL~\cite{fan2025ssrl} & Adapt-Search & \cmark & \makecell{Simulated\\Self-Search} & GRPO & Rule-based ORM & \makecell[l]{Format \\ Answer EM} & Single-agent & \cite{kwiatkowski2019natural,joshi2017triviaqa,yang2018hotpotqa,trivedi2022musique,ho2020constructing,press2022measuring}\\
\midrule
Pangu DeepDiver~\cite{shi2025pangu} & \makecell[l]{Adapt-Search\\Search Intensity} & \cmark & Real-world & GRPO & Rule-based ORM & \makecell[l]{Format\\Answer EM\\Extra Search} & Single-agent & \cite{shi2025pangu,press2022measuring,wei2024measuring,krishna2024frames}\\
\midrule
ReZero~\cite{dao2025rezero} & \makecell[l]{Search Intensity} & \xmark & Real-world & GRPO & ORM+PRM & \makecell[l]{Format\\Answer LLM-Judge\\Retry} & Step-level & \cite{menlo2025apollo3}\\
\midrule
StepSearch~\cite{wang2025stepsearch} & \makecell[l]{Adapt-Search\\Search Intensity} & \xmark & Real-world & PPO & Rule-based ORM+PRM & \makecell[l]{Format\\Answer F1\\Search Key\\Information Gain\\Redundancy Penalty} & Step-level & \cite{yang2018hotpotqa,trivedi2022musique,ho2020constructing,press2022measuring}\\
\midrule

VERITAS~\cite{xu2025VERITAS} & \makecell[l]{Adapt-Search\\R-Aware Opt.} & \xmark & Real-world & PPO & ORM+PRM & \makecell[l]{Answer EM\\Enhancing Faithfulness} & Step-level & \cite{kwiatkowski2019natural,joshi2017triviaqa,mallen2022not,yang2018hotpotqa,trivedi2022musique,ho2020constructing,press2022measuring}\\
\midrule

ReasonRAG~\cite{zhang2025process} & \makecell[l]{Search Efficiency\\R-S Inter.} & \xmark & \makecell[l]{Real-world\\MCTS} & DPO & PRM & \makecell[l]{Shortest Path} & Step-level & \cite{mallen2022not,yang2018hotpotqa,ho2020constructing,press2022measuring,trivedi2022musique} \\
\midrule
Web-Sailor~\cite{li2025websailor} & \makecell[l]{Adapt-Search\\Ctx-Mem.} & \cmark & Real-world & DUPO & ORM & \makecell[l]{Format\\Answer F1} & Single-agent &  \cite{li2025sailorfogqa,wei2025browsecomp,zhou2025browsecomp,mialon2023gaia,xbench} \\
\midrule
WebSailor-V2~\cite{li2025websailorv2} & \makecell[l]{Multi-tool\\Ctx-Mem.} & \cmark & Real-world & GRPO & Rule-based ORM & \makecell[l]{Format\\Answer F1} & Single-agent & \cite{li2025sailorfogqav2,wei2025browsecomp,zhou2025browsecomp,mialon2023gaia,xbench,phan2025humanity,du2025deepresearch}\\
\midrule
Search Wisely~\cite{wu2025search} & Search Efficiency & \xmark & Real-world & $\beta$-GRPO & Rule-based ORM & Confidence-based Answer EM & Single-agent & \cite{kwiatkowski2019natural,yang2018hotpotqa,joshi2017triviaqa,ho2020constructing,press2022measuring,trivedi2022musique}\\
\midrule
ZeroSearch~\cite{sun2025zerosearch} & Search Efficiency & \cmark & \makecell[l]{Simulated\\Curriculum} & \makecell[l]{PPO\\GRPO\\Reinforce} & Rule-based ORM & Answer F1 & Single-agent & \cite{kwiatkowski2019natural,joshi2017triviaqa,mallen2022not,yang2018hotpotqa,ho2020constructing,trivedi2022musique,press2022measuring}\\
\midrule
ParallelSearch~\cite{zhao2025parallelsearch} & Search Efficiency & \cmark & \makecell[l]{Real-world} & GRPO & Rule-based ORM & \makecell[l]{Format \\ Answer EM\\Query Decomopse\\Search count} & Single-agent & \cite{kwiatkowski2019natural,joshi2017triviaqa,mallen2022not,yang2018hotpotqa,ho2020constructing,trivedi2022musique,press2022measuring}\\
\midrule
RAG-R1~\cite{tan2025ragr1} & \makecell[l]{Search Efficiency\\
Conv-Reform.}& \cmark & Real-world & PPO & ORM & Answer EM & Single-agent & \cite{kwiatkowski2019natural,mallen2022not,joshi2017triviaqa,yang2018hotpotqa,ho2020constructing,trivedi2022musique,press2022measuring}\\
\midrule
ConvSearch-R1~\cite{zhu2025convsearch} &Conv-Reform. & \cmark & Real-world & GRPO & ORM & \makecell[l]{Format\\Rank-Incentive} & Step-level & \cite{adlakha2022topiocqa,anantha2021open}\\
\midrule
MaskSearch~\cite{wu2025masksearch} & \makecell[l]{Conver. Reform.\\R–S Inter.} & \cmark & \makecell[l]{Real-world\\Curriculum\\RAMP} & DAPO & Rule-based ORM & \makecell[l]{Format\\Answer Recall\\Length penalty} & Single-agent & \cite{yang2018hotpotqa,zhu2024fanoutqa,trivedi2022musique,ho2020constructing,press2022measuring,vu2024freshqa} \\
\midrule
DeepRetrieval~\cite{jiang2025deepretrieval} & R-Aware Opt. & \cmark & Simulated & PPO & ORM & \makecell[l]{Format\\Answer Recall} & Single-level & \cite{kwiatkowski2019natural,joshi2017triviaqa,rajpurkar2016squad,thorne2018fever,wadden2020fact}\\
\midrule
WebThinker~\cite{li2025webthinker} & Search Efficiency & \xmark & Real-world & DPO & PRM & \makecell[l]{Answer EM\\Tool Calls\\Length penalty} & Single-agent & \cite{rein2024gpqa,mialon2023gaia,wu2025webwalker,phan2025humanity,du2025supergpqa,openthoughts2025open,yuan2025naturalreasoning,li2024numinamath} \\
\midrule
s3~\cite{jiang2025s3} & R-Aware Opt. & \cmark & Simulated & PPO & Rule-based ORM & Gain Beyond RAG & Module-level & \cite{kwiatkowski2019natural,joshi2017triviaqa,mallen2022not,yang2018hotpotqa,ho2020constructing,trivedi2022musique}\\
\midrule
R-Search~\cite{zhao2025r} & R–S Inter. & \xmark & Real-world & \makecell[l]{PPO\\GRPO} & Rule-based ORM+PRM & \makecell[l]{Format\\Answer F1\\Evidence Quality} & Single-agent & \cite{yang2018hotpotqa,ho2020constructing,trivedi2022musique,press2022measuring}\\
\midrule
AutoRefine~\cite{shi2025search} & R–S Inter. & \xmark & Real-world & GRPO & ORM plus PRM & \makecell[l]{Answer F1\\Retrieval Reward} & Step-level & \cite{kwiatkowski2019natural,joshi2017triviaqa,yang2018hotpotqa,ho2020constructing,trivedi2022musique,press2022measuring,mallen2022not}\\
\midrule
EvolveSearch~\cite{zhang2025evolvesearch} & R–S Inter. & \cmark & \makecell[l]{Real-world\\Self-evolving} & GRPO & ORM & \makecell[l]{Format\\Answer LLM-Judge\\} & Single-agent & \cite{kwiatkowski2019natural,joshi2017triviaqa,yang2018hotpotqa,ho2020constructing,trivedi2022musique,press2022measuring,mallen2022not}\\
\midrule
O$^2$-Searcher~\cite{mei2025o2} & R–S Inter. & \cmark & Simulated & GRPO & Rule-based ORM & \makecell[l]{Format\\Diversity reward\\Factual reward} & Single-agent & \cite{kwiatkowski2019natural,yang2018hotpotqa,joshi2017triviaqa,mallen2022not,ho2020constructing,trivedi2022musique,press2022measuring}\\
\midrule
Atom-Searcher~\cite{deng2025atom} & \makecell[l]{R–S Inter.} & \cmark & \makecell[l]{Real-world\\Curriculum} & GRPO & 
PRM+Rule-based ORM & \makecell[l]{Format\\Answer F1\\Atomic thought reward} & Step-level & \cite{kwiatkowski2019natural,joshi2017triviaqa,yang2018hotpotqa,ho2020constructing,trivedi2022musique,press2022measuring,mallen2022not}\\
\midrule
ReSum~\cite{wu2025resum} & Ctx-Mem. & \xmark & \makecell[l]{Real-world\\} & Resume-GRPO & ORM & Answer LLM-Judge & Single-agent & \cite{mialon2023gaia,wei2025browsecomp,zhou2025browsecomp,wei2024measuring,wu2025webwalker,xbench}\\
\midrule
SFR-DeepResearch~\cite{nguyen2025sfr} & \makecell[l]{Ctx-Mem.\\Multi-tool} & \xmark & Real-world & REINFORCE & ORM & Answer LLM-Judge & Single-agent & \cite{krishna2024frames,mialon2023gaia,phan2025humanity}\\
\midrule
MAO-ARAG~\cite{chen2025mao} & P–E Orches. & \cmark & Real-world & PPO & ORM & \makecell[l]{Format\\Cost Penalty\\Answer F1} & Multi-agent & \cite{kwiatkowski2019natural,yang2018hotpotqa,ho2020constructing,trivedi2022musique,press2022measuring,mallen2022not,min2020ambigqa}\\
\midrule
OPERA~\cite{liu2025opera} & P–E Orches. & \cmark & Real-world & MAPGRPO & PRM+ORM & \makecell[l]{Answerer Reward\\Planer Reward\\Rewriter  Reward} & Multi-agent & \cite{yang2018hotpotqa,ho2020constructing,trivedi2022musique,kwiatkowski2019natural,tang2024multihop}\\
\midrule
AI-SearchPlanner~\cite{mei2025ai} & P–E Orches. & \xmark & Real-world & PPO & ORM & \makecell[l]{Answer LLM-Judge\\Trajectory Rationality} & Module-level & \cite{kwiatkowski2019natural,joshi2017triviaqa,yang2018hotpotqa,ho2020constructing,trivedi2022musique,press2022measuring,mallen2022not}\\
\midrule
SIRAG~\cite{wang2025sirag} & Cooperative & \xmark & Real-world & PPO & PRM & Process LLM-Judge & Multi-agent & \cite{ho2020constructing,yang2018hotpotqa,kwiatkowski2019natural,mallen2022not}\\
\midrule
MMOA-RAG~\cite{chen2025improving} & \makecell[l]{Cooperative\\R-aware Opt.}& \xmark & Real-world & MA-PPO & Rule-based ORM & \makecell[l]{Answer F1\\Efficiency penalty} & Multi-agent & \cite{yang2018hotpotqa,ho2020constructing,min2020ambigqa}\\
\midrule
Tool-Star~\cite{dong2025tool} & Multi tool & \cmark & Real-world & \makecell[l]{REINFORCE++\\GRPO\\DPO} & Rule-based ORM & \makecell[l]{Format\\Answer EM} & Single-agent & \cite{lightman2023let,hendrycks2021measuring,cobbe2021training,wu2025webwalker,yang2018hotpotqa,ho2020constructing,trivedi2022musique,press2022measuring}\\
\midrule
WebWatcher~\cite{geng2025webwatcher} & \makecell[l]{Multi tool\\Multi-modal} & \cmark & Real-world & GRPO & ORM & \makecell[l]{Format\\Answer LLM-Judge} & Single-agent & \cite{phan2025humanity,li2025mm,fu2025livevqa,jiang2024mmsearch,cheng2025simplevqa}\\
\midrule
Visual-ARFT~\cite{liu2025visual} & \makecell[l]{Multi-modal\\Multi-tool\\Adapt-Search} & \cmark & Real-world & GRPO & Rule-based ORM+PRM & \makecell[l]{Format\\Answer F1\\Query Semantic Sim.} & Single-agent & \cite{liu2025matsearch}\\
\midrule
VRAG-RL~\cite{wang2025vrag} & \makecell[l]{Multi-modal\\Search Efficiency} & \cmark & Simulated & GRPO & ORM & \makecell[l]{Format\\Answer LLM-Judge\\Retrieval Efficiency} & Single-agent & \cite{tanaka2023slidevqa,wang2025vidorag,ma2024mmlongbench}\\
\midrule
MMSearch-R1~\cite{wu2025mmsearch} & \makecell[l]{Multi-modal \\Search Efficiency}& \xmark & Real-world & GRPO & Rule-based ORM & \makecell[l]{Format\\Answer EM\\Search Penalty} & Single-agent & \cite{wu2025mmsearch,chen2023can,jiang2024mmsearch,cheng2025simplevqa,fu2025livevqa}\\
\midrule
GRAIL~\cite{chang2025grail} & \makecell[l]{Adapt-Search\\Struct-Nav.} & \cmark & \makecell[l]{Real-world\\Graph Env.} & GRPO & PRM & Process LLM-Judge & Single-agent & \cite{berant2013webquestions,puerto2021metaqa,talmor2018web}\\
\midrule
DynaSearcher~\cite{hao2025dynasearcher} & Struct-Nav. & \xmark & \makecell[l]{Real-world\\Graph Env.\\KG+Doc Search} & GRPO & Rule-based ORM & \makecell[l]{Format\\Answer F1\\Information Gain\\Retrieval Penalty} & Single-agent & \cite{yang2018hotpotqa,ho2020constructing,trivedi2022musique,press2022measuring,krishna2024frames}\\
\midrule
HARIS~\cite{hu2025coordinating} & R-S Inter. & \xmark & Real-world & GRPO & Rule-based ORM & \makecell[l]{Format\\Answer Accuracy\\Decision Accuracy} & Multi-agent & \cite{ma2024ex,jiang2020hover,si2024checkwhy} \\
\midrule
DeepNote~\cite{wang2024retriever} & \makecell[l]{Adapt-Search\\Conv-Reform.} & \xmark & Real-world & DPO & - & - & Single-agent & \cite{yang2018hotpotqa,ho2020constructing,trivedi2022musique,geva2021strategyqa,stelmakh2022asqa} \\
\midrule
DeepResearcher~\cite{zheng2025deepresearcher} & \makecell[l]{Adapt-Search\\Search Efficiency\\Ctx-Mem.} & \xmark & Real-world & GRPO & Rule-based ORM & \makecell[l]{Format\\Answer F1} & Module-level &\cite{kwiatkowski2019natural,joshi2017triviaqa,yang2018hotpotqa,ho2020constructing}\\
\midrule
SWiRL~\cite{goldie2025synthetic} & \makecell[l]{Adapt-Search\\R-S Inter.} & \cmark & Real-world & PPO & PRM & \makecell[l]{Step LLM-Judge} & Step-level & \cite{yang2018hotpotqa,trivedi2022musique,wu2024cofca,cobbe2021training,qi2021answering} \\
\midrule
WebDancer~\cite{wu2025webdancer} & \makecell[l]{Multi tool} & \cmark & Real-world & DAPO & ORM & \makecell[l]{Answer EM} & Single-agent & \cite{mialon2023gaia,wu2025webwalker,zhou2025browsecomp,wei2025browsecomp} \\
\midrule
MedResearcher-R1~\cite{yu2025medresearcher} & \makecell[l]{Adpt-Search\\Multi-Tool} & \cmark &  \makecell[l]{Real-world\\Medical Tool} & GRPO & ORM & \makecell[l]{Answer Acc\\Response Quality\\Efficiency penalty} & Single-agent & \cite{xbench,mialon2023gaia,chen2025medbrowsecomp} \\
\midrule
Lucy~\cite{dao2025lucy} & \makecell[l]{Search Efficiency\\R–S Inter.} & \cmark & \makecell[l]{Real-world\\SLMs} & DAPO & Rule-based ORM & \makecell[l]{Format/XML validity\\Answer EM\\Tool exec. success\\Visit/Search ratio\\Efficient thinking} & Single-agent & \cite{wei2024measuring}\\
\midrule
ASearcher~\cite{gao2025beyond}& \makecell[l]{R-S Inter.\\Ctx-Mem.\\Multi-tool} & \cmark & \makecell[l]{Real-world\\Browser Env.\\Asynchronous} & GRPO & ORM & Answer LLM-Judge & Single-agent & \cite{kwiatkowski2019natural,joshi2017triviaqa,mallen2022not,yang2018hotpotqa,ho2020constructing,trivedi2022musique,press2022measuring, krishna2024frames,mialon2023gaia,xbench}\\
\midrule
WebExplorer~\cite{liu2025webexplorer} & \makecell[l]{Ctx-Mem.\\Conv-Reform.} & \cmark & \makecell[l]{Real-world\\Curriculum} & GRPO & Rule-based ORM & \makecell[l]{Format\\Answer EM} & Single-agent & \cite{zhou2025browsecomp,wei2025browsecomp,mialon2023gaia,wu2025webwalker,krishna2024frames,xbench,phan2025humanity} \\
\midrule
WebResearcher~\cite{qiao2025webresearcher} & Multi-tool & \cmark & \makecell[l]{Real-world\\Curriculum} & GSPO & Rule-based ORM & Answer EM & Single-agent & \cite{phan2025humanity,mialon2023gaia,wei2025browsecomp,zhou2025browsecomp,xbench,krishna2024frames} \\
\midrule
RECON~\cite{xu2025reconreasoningcondensationefficient} & Ctx-Mem. & \cmark & \makecell[l]{Real-world} & PPO & Rule-based ORM & Answer EM & Single-agent & \cite{kwiatkowski2019natural,joshi2017triviaqa,press2022measuring,yang2018hotpotqa,ho2020constructing,trivedi2022musique,mallen2022not} \\
\midrule
AgentGym-RL~\cite{xi2025agentgym} & \makecell[l]{Cooperative\\Multi tool} & - & - & - & - & - & Unified RL Agentic Framework & -\\
\midrule
Chain-of-Agents~\cite{li2025chain} & \makecell[l]{Cooperative\\Multi tool} & - & - & - & - & - & Unified RL Agentic Framework & -\\
\midrule
Verl~\cite{sheng2024hybridflow} & \makecell[l]{Multi tool} & - & - & - & - & - & Unified RL Agentic Framework & - \\
\midrule
VerlTool~\cite{jiang2025verltool} & \makecell[l]{Multi tool} & - & - & - & - & - & Unified RL Agentic Framework & -\\
\bottomrule
\end{longtable}
}
\end{center}